\documentclass[10pt,twocolumn,letterpaper]{article}
\usepackage[pagenumbers]{iccv}

\definecolor{rank1}{RGB}{223,101,97}
\definecolor{rank2}{RGB}{233,161,162}
\definecolor{rank3}{RGB}{251,205,185}
\usepackage{colortbl}
\usepackage{xcolor}
\usepackage{multirow}
\usepackage{bm}
\usepackage{stfloats}
\usepackage{algorithm,algorithmic}

\definecolor{iccvblue}{rgb}{0.21,0.49,0.74}
\usepackage[pagebackref,breaklinks,colorlinks,allcolors=iccvblue]{hyperref}
\usepackage[belowskip=-10pt,aboveskip=0pt]{caption}

\title{Dynamics-Aware Gaussian Splatting Streaming \\ Towards Fast On-the-Fly 4D Reconstruction}

\author{Zhening Liu$^{1}$, Yingdong Hu$^1$, Xinjie Zhang$^{1}$, Rui Song$^{1}$, Jiawei Shao$^{1,2}$, Zehong Lin$^{1}$\thanks{Corresponding author}~~, Jun Zhang$^1$\\
$^{1}$Hong Kong University of Science and Technology,\\
$^2$ Institute of Artificial Intelligence (TeleAI), China Telecom \\
{\tt\small \{zhening.liu,yhudj,xinjie.zhang,rrui.song\}@connect.ust.hk,} \\
{\tt\small shaojw2@chinatelecom.cn,}
{\tt\small\{eezhlin,eejzhang\}@ust.hk}
}

\begin{document}
% \maketitle
\twocolumn[{
\renewcommand\twocolumn[1][]{#1}
\maketitle
\begin{center}
    \captionsetup{type=figure}
    \includegraphics[width=0.9\linewidth]{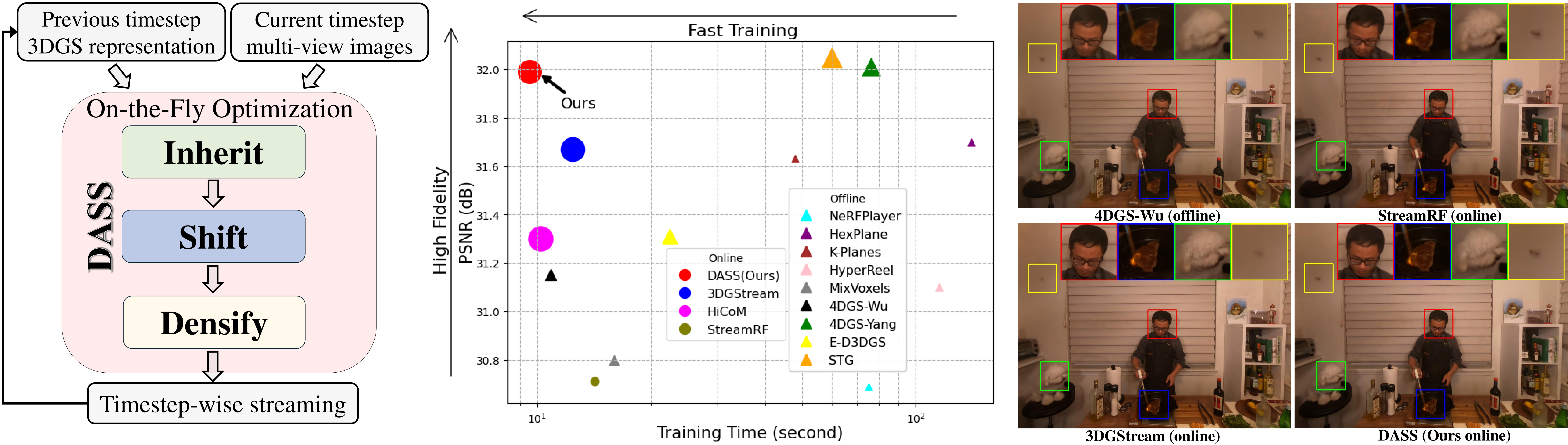}
    \captionof{figure}{(Left) An overview of our three-stage pipeline, referred to as \textit{DASS}. (Middle) Performance of 4D dynamic spatial reconstruction methods in terms of per-timestep training time and reconstruction quality (PSNR) on the N3DV dataset~\cite{li2022neural}. Online methods are represented by circles, while offline methods are indicated by triangles. The size of each marker is proportional to the rendering speed (FPS). Our method achieves the fastest online training speed, superior reconstruction quality, and real-time rendering capability. (Right) Visual comparisons of baseline methods, illustrating that our method preserves fine-grained details and recovers diverse dynamics.}
    \vspace{11pt}
    \label{fig:teaser}
\end{center}
}]

\begin{abstract}
The recent development of 3D Gaussian Splatting (3DGS) has led to great interest in 4D dynamic spatial reconstruction. 
Existing approaches mainly rely on full-length multi-view videos, while there has been limited exploration of online reconstruction methods that enable on-the-fly training and per-timestep streaming. 
Current 3DGS-based streaming methods treat the Gaussian primitives uniformly and constantly renew the densified Gaussians, thereby overlooking the difference between dynamic and static features as well as neglecting the temporal continuity in the scene.
To address these limitations, we propose a novel three-stage pipeline for iterative streamable 4D dynamic spatial reconstruction. Our pipeline comprises a selective inheritance stage to preserve temporal continuity, a dynamics-aware shift stage to distinguish dynamic and static primitives and optimize their movements, and an error-guided densification stage to accommodate emerging objects. 
Our method achieves state-of-the-art performance in online 4D reconstruction, demonstrating the fastest on-the-fly training, superior representation quality, and real-time rendering capability.
Project page: \url{https://www.liuzhening.top/DASS}
\end{abstract}

\section{Introduction}
\label{sec:intro}
The rapid advancements of stereoscopic cameras and rendering techniques have expanded human visual perception from 2D planes to spatial 3D representations. This evolution has paved the way for 4D dynamic free-viewpoint video (FVV) reconstruction by integrating the temporal dimension, which unlocks substantial potential for a wide range of applications, including augmented/virtual reality (AR/VR)~\cite{arena2022overview} and holographic communications~\cite{tu2024tele}. Nevertheless, constructing 4D dynamic FVVs from multi-view 2D inputs remains a significant challenge. 

In recent years, Neural Radiance Field (NeRF)~\cite{nerf} has emerged as a promising approach for spatial representation from multi-view inputs, which optimizes neural networks to estimate color and density based on spatial positions and viewpoints.
The extensions of NeRF to dynamic scene reconstruction have demonstrated remarkable effectiveness~\cite{li2022neural,DNERF,song2023nerfplayer,fridovich2023k,guo2024motion}, yielding photo-realistic novel view synthesis. However, the efficiency of NeRF-based methods is severely hindered by their low rendering speed due to the dense queries of neural networks. To address this issue, 3D Gaussian Splatting (3DGS)~\cite{kerbl20233d} has been proposed as a solution to provide high-quality reconstruction and real-time rendering capabilities, leveraging its flexible point-based primitive design and tile-based differentiable rasterization. Subsequent efforts~\cite{katsumata2024compact,yangreal,li2024spacetime,wu20244d,bae2024per} have been dedicated to applying Gaussian Splatting for 4D dynamic reconstruction, with representative works integrating the time dimension into each Gaussian primitive~\cite{yangreal} or learning spatio-temporal deformations~\cite{li2024spacetime,wu20244d}.

Despite these advancements, most NeRF-based and 3DGS-based methods for 4D dynamic spatial reconstruction rely on full-length multi-view video inputs, which are non-causal in nature. This reliance limits their applicability in scenarios such as live streaming, where only per-timestep causal inputs are available, necessitating on-the-fly training.
This problem can be formalized as the iterative reconstruction of 3D space at the current timestep, leveraging the reconstruction caches from the previous timestep and the multi-view inputs of the current timestep. The key challenges in this context are two-fold: (i) \textit{how to effectively model temporal variations in 3D space across consecutive timesteps} and (ii) \textit{how to facilitate efficient optimization convergence from the previous timestep to the current one}. Critical performance metrics for this problem include both the quality of novel view synthesis and the time efficiency.

One intuitive solution is to optimize a new set of 3DGS primitives for each timestep independently. However, this strategy incurs substantial computational and storage costs, as it requires tuning and storing all 3DGS parameters for every timestep. A representative baseline, 3DGStream~\cite{sun20243dgstream}, addresses this issue by optimizing the transformation of Gaussian positions and rotation quaternions while adaptively densifying new Gaussians. Although this method achieves fast and high-quality reconstruction, it treats the entire scene uniformly and fails to distinguish between inherent dynamics and statics in the scene. In practice, dynamic and static components have different deformation characteristics.
For instance, moving objects, such as humans or animals, display substantial dynamics, with the Gaussian properties like positions experiencing significant offsets. In contrast, static background and stationary objects show minimal movement, where Gaussians remain unchanged or undergo slight jitters. 
Consequently, uniformly modeling the transformation of all Gaussians serves as a sub-optimal strategy. Moreover, renewing the added Gaussian primitives for each timestep overlooks the temporal continuity inherent in the scene, leading to inefficiencies in both computation and representation.

Based on these insights, we propose a dynamics-aware 3DGS streaming paradigm for on-the-fly 4D reconstruction, termed \textit{DASS}, where the optimization of each timestep comprises three stages: inheritance, shift, and densification. Specifically, considering the temporal continuity, the newly densified Gaussians in the previous timestep are likely to persist in subsequent timesteps. Therefore, instead of renewing and optimizing added Gaussians for each timestep from scratch, we propose a selective inheritance mechanism to adaptively include a portion of the added Gaussians from the previous timestep using a learnable selection mask. Then, in the shift stage, we employ 2D dynamics-related priors, optical flow~\cite{xu2023unifying} and Gaussian segmentation~\cite{ye2025gaussian} to calculate a per-Gaussian dynamics mask. Subsequently, we assign deformation layers to learn the offsets of dynamic and static Gaussians with different representation complexities, boosting a more dedicated optimization. In the densification stage, apart from the Gaussian offsets that present the deformations of existing objects, new Gaussian primitives are introduced to accommodate newly emerging objects. In this stage, both positional gradients and error maps from the shift stage serve as criteria for identifying regions that require densification. Iteratively, the inheritance stage in the subsequent timestep will process these added Gaussians, thereby mitigating errors in the shift stage and reducing the optimization burden in the densification stage. 
Our three-stage pipeline effectively captures dynamic spatial components and exploits the temporal correlation, achieving fast and high-fidelity on-the-fly 4D reconstruction. 
Our main contributions are summarized as follows:
\begin{itemize}
    \item We propose a novel three-stage pipeline for 4D dynamic spatial reconstruction that supports on-the-fly training and per-timestep streaming. Our method builds on causal inputs and eliminates the need for full-length multi-view videos, thereby enhancing the practicability.
    \item Our approach seamlessly integrates three stages to optimize reconstruction quality. By selectively inheriting Gaussians from the preceding timestep, effectively distinguishing dynamic primitives to allocate optimization emphasis, and refining under-reconstructed areas using gradient information and optimization errors, our method ensures high-fidelity dynamic spatial reconstruction.
    \item Extensive experiments demonstrate the superiority of our method in multiple aspects, including the highest online training speed, superior reconstruction quality, and real-time rendering capability.
\end{itemize}

%-------------------------------------------------------------------------
\section{Related Works}
\label{sec:related}
\subsection{Neural Static Scene Representation}
In recent years, reconstructing 3D representations from 2D plane visual inputs has experienced significant advancements, driven by the development of NeRF~\cite{nerf} and 3DGS~\cite{kerbl20233d}. NeRF-based methods represent spatial scenes by optimizing multi-layer perceptrons (MLPs) and generate novel views through volume rendering~\cite{kajiya1984ray}. Subsequent research has enhanced both the training and rendering efficiency through grid-based designs~\cite{muller2022instant,xu2022point,fridovich2022plenoxels,fridovich2023k}. Nonetheless, NeRF-based approaches typically require dense ray tracing and struggle to fulfill high-speed rendering. Recently, 3DGS~\cite{kerbl20233d} has emerged to address these limitations by utilizing explicit unstructured scene representation while preserving point-based differentiable splatting rendering~\cite{zwicker2001ewa}. This approach achieves real-time rendering speed and photo-realistic quality. 
Based on these advancements, subsequent studies have further enhanced the representation efficiency~\cite{lu2024scaffold,chen2024hac,liu2024compgs,fan2023lightgaussian,lee2024compact,zhang2025gaussianimage,fan2024instantsplat,ye20253d} and expanded applications in understanding and editing~\cite{ye2025gaussian,zhou2024feature,chen2024gaussianeditor,shen2025flashsplat,wang2024gscream}.

\subsection{Neural Dynamic Scene Reconstruction}
Extending static scene representation to dynamic FVV reconstruction remains a significant challenge, primarily due to the difficulties in modeling temporal correlations and variations in 3D space. 
To address this issue, several studies have extended NeRF to incorporate spatio-temporal structures~\cite{li2022streaming,cao2023hexplane,song2023nerfplayer,fridovich2023k,wang2023mixed}, facilitating dynamic space reconstruction. Similarly, dynamic scene reconstruction methods based on Gaussian Splatting have been proposed, which can be categorized into deformation-based methods, 4D primitive-based methods, and iterative streaming methods. Deformation-based methods~\cite{wu20244d,lu20243d,yang2024deformable,bae2024per} maintain 3D Gaussian representations and optimize neural deformable fields to capture temporal variations. 4D primitive-based methods~\cite{yangreal,li2024spacetime,zhang2024mega} augment Gaussian primitives by integrating the temporal axis as an intrinsic property, thereby directly learning the spatio-temporal distributions. Note that both categories of methods rely on full-length multi-view videos for training, which limits their ability to support on-the-fly training and per-timestep streaming. In contrast, iterative streaming methods build upon previously converged representations and perform per-timestep optimization for each timestep's multi-view inputs, thereby addressing the above limitations. However, this area remains largely underexplored. Our work aims to develop an iterative streaming pipeline that accelerates per-timestep optimization convergence and enhances the reconstruction quality. 

The work most related to ours is 3DGStream~\cite{sun20243dgstream}. It proposes a two-stage optimization process for each timestep, where the first stage optimizes a grid-based MLP~\cite{muller2022instant} for Gaussian property transformation and the second stage wisely adds new Gaussians based on positional gradients. However, 3DGStream does not consider the intrinsic difference between dynamic and static features in the scene and discards the added primitives at each timestep, thereby limiting its ability to leverage temporal continuity across timesteps. 
Concurrently, other studies on iterative 4D reconstruction focus on different aspects, such as reducing streaming storage overhead on mobile devices~\cite{wang2024v} and exploring integration into communication systems~\cite{liu2024swings,sun2024multi}.
In contrast, our work focuses on achieving fast convergence in per-timestep optimization while maintaining high reconstruction quality, which is orthogonal to the contributions of aforementioned studies.

\begin{figure*}[t]
    \centering
    \includegraphics[width=0.9\linewidth]{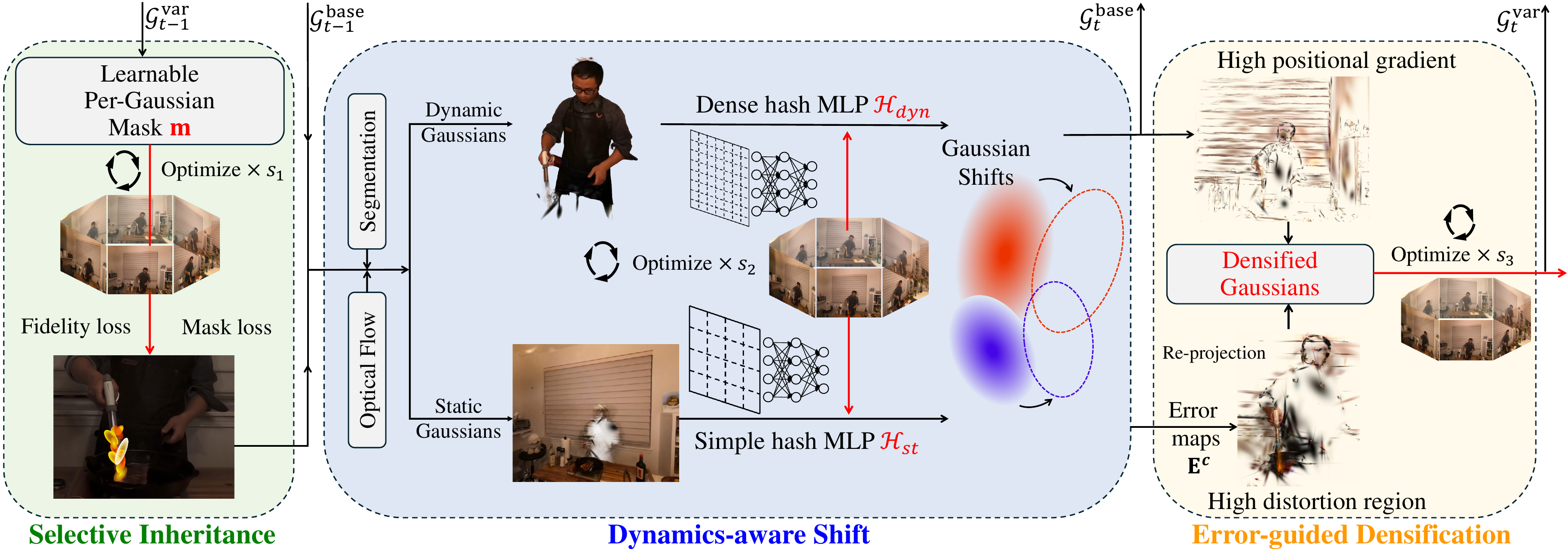}
    \caption{Overview of our proposed DASS framework. The selective inheritance stage (Green) exploits the temporal continuity and adaptively preserves Gaussians from the previous timestep. The dynamics-aware shift stage (Blue) distinguishes the dynamic and static elements and optimizes the deformations. The error-guided densification stage (Yellow) detects and densifies the areas with weak reconstruction based on positional gradients and distortions. Variables highlighted in red represent learnable parameters in each stage, whose training is significantly lightweight compared to tuning all Gaussian parameters.}
    \label{fig:pipeline}
\end{figure*}

\vspace{-0.4em}
\section{Methodology}
\label{sec:method}
\subsection{Overview}
Our work addresses the problem of Gaussian Splatting streaming, which is formalized as an iterative 4D dynamic spatial reconstruction process, enabling on-the-fly training and per-timestep streaming.
At each timestep $t$, the optimization begins with the representation from the previous timestep, denoted by $\bm{\mathcal{G}}_{t-1}$, and derives the current representation $\bm{\mathcal{G}}_{t}$, using multi-view inputs from $C$ viewpoints.

Our streaming pipeline, referred to as \textit{DASS}, represents the 3D space at each timestep $t$ using two sets of Gaussian primitives: quantity-consistent base Gaussians and temporally quantity-varied densified Gaussians, i.e., $\bm{\mathcal{G}}_{t}:=\{\bm{\mathcal{G}}_t^{\textrm{base}},\bm{\mathcal{G}}_t^{\textrm{var}}\}$. The base Gaussians $\bm{\mathcal{G}}^{\textrm{base}}$, initialized from the zero timestep representation $\bm{\mathcal{G}}_{0}$, constitute the fundamental elements shared across all timesteps and maintain a fixed number of $N$ Gaussian primitives throughout the streaming process. In contrast, the number of Gaussians in $\bm{\mathcal{G}}^{\textrm{var}}$ varies across timesteps, capturing per-timestep unique emerging objects that are not represented in $\bm{\mathcal{G}}^{\textrm{base}}$.

Fig. \ref{fig:pipeline} depicts an overview of our framework, where the optimization for each timestep comprises three stages: inheritance, shift, and densification. 
The selective inheritance stage (Sec. \ref{sec:inherit}) preserves temporal consistency by incorporating all base Gaussians $\bm{\mathcal{G}}_{t-1}^{\textrm{base}}$ from the previous timestep, while selectively inheriting the temporally-varied Gaussians $\bm{\mathcal{G}}_{t-1}^{\textrm{var}}$ using a per-Gaussian learnable mask. This mask effectively eliminates redundant Gaussians while retaining those essential for the current timestep.
Then, the shift stage (Sec. \ref{sec:shift}) transforms the inherited Gaussians to align with the current timestep, capturing the movements of existing elements. In this stage, a Gaussian-level dynamics mask is estimated to classify the Gaussians into dynamic and static groups, which are then processed by two hash-encoding MLPs with different complexities. This shift stage efficiently manages the movements and rotations of Gaussians, accommodating changes from the previous timestep.
In the densification stage (Sec. \ref{sec:densi}), emerging objects, such as coffee being poured from a cup or flames emanating from an oven, are identified by analyzing positional gradients and distortions resulting from the shift stage. This stage densifies a limited number of additional Gaussians to accurately represent these emerging objects. 
Our proposed design facilitates fast per-timestep convergence and achieves high-fidelity novel view synthesis. The details of each stage are elaborated in the following subsections.

\subsection{Selective Inheritance}
\label{sec:inherit}
Exploiting temporal correlations is critical for effective and efficient Gaussian Splatting streaming. Due to the strong temporal consistency inherent in the scene, the base Gaussians $\bm{\mathcal{G}}^{\textrm{base}}$ and a subset of the temporally-varied Gaussians $\bm{\mathcal{G}}^{\textrm{var}}$ from the previous timestep remain applicable for subsequent timesteps, since they have been specifically optimized for the scene. Leveraging these timestep-wise priors and dependencies can significantly accelerate on-the-fly training convergence. 
However, previous methods, such as 3DGStream~\cite{sun20243dgstream}, typically retain the base Gaussians $\bm{\mathcal{G}}_{t-1}^{\textrm{base}}$ while discarding the previously densified Gaussians $\bm{\mathcal{G}}_{t-1}^{\textrm{var}}$ to prevent excessive accumulation of Gaussians, which can lead to prohibitive training and streaming overhead. To address this limitation, we propose a selective inheritance mechanism for $\bm{\mathcal{G}}_{t-1}^{\textrm{var}}$. Our approach adaptively selects a subset of the Gaussians that are beneficial for the current reconstruction while controlling the total number of Gaussians to avoid excessive accumulation. 

Specifically, for each Gaussian primitive in  $\bm{\mathcal{G}}^{\textrm{var}}_{t-1}$, we assign a learnable parameter, resulting in a parameter vector $\mathbf{m}\in\mathbb{R}^{N\times 1}$. We then apply the sigmoid function to $\mathbf{m}$, yielding $\textrm{sigmoid}(\mathbf{m})$, followed by quantization to generate a binary mask. This mask indicates the selection of inherited Gaussians and is element-wise multiplied with the Gaussian opacities and scales as follows:
\begin{equation}
    \setlength\abovedisplayskip{3pt}
    \setlength\belowdisplayskip{2pt}
    \begin{aligned}
    \mathbf{o}_r&=\textrm{Quant}(\textrm{sigmoid}(\mathbf{m}))\circ \mathbf{o}, \\
    \mathbf{s}_r&=\textrm{Quant}(\textrm{sigmoid}(\mathbf{m}))\circ \mathbf{s},
\end{aligned}
\end{equation}
where $\mathbf{o}_r$ and $\mathbf{s}_r$ are the opacity and scale values used during rendering, $\mathbf{o}$ and $\mathbf{s}$ are the original opacity and scale values, and $\circ$ represents the element-wise multiplication. When the mask is quantized to zero, the corresponding Gaussian has zero opacity and scale, thus contributing nothing to the rendering process. Therefore, this mask determines whether each Gaussian is retained for rendering. By optimizing $\mathbf{m}$, the method selectively inherits the most relevant densified Gaussians from $\bm{\mathcal{G}}^{\textrm{var}}_{t-1}$. The loss function for optimizing $\mathbf{m}$ is expressed as follows:
\begin{equation}
    \setlength\abovedisplayskip{1pt}
    \setlength\belowdisplayskip{1pt}
    \mathcal{L}=(1-\lambda)\mathcal{L}_1+\lambda\mathcal{L}_{\textrm{D-SSIM}}+\lambda_{\textrm{inher}}\sum\textrm{sigmoid}(\mathbf{m}).
\end{equation}
The first two terms represent the fidelity loss in the vanilla 3DGS~\cite{kerbl20233d}, comprising pixel-level $L_1$ loss and D-SSIM~\cite{wang2003multiscale} loss, respectively. The final term, weighted by $\lambda_{\textrm{inher}}$, is referred to as the mask loss and acts as a regularizer that encourages the parameters in $\mathbf{m}$ to approach zero. This regularization effectively reduces the number of inherited Gaussians and controls numerical accumulation, while the fidelity loss aims to retain Gaussians that are beneficial for reconstruction. Consequently, with only a few optimization iterations, this trade-off effectively optimizes the learnable $\mathbf{m}$ and selectively inherits important Gaussians from $\bm{\mathcal{G}}^{\textrm{var}}_{t-1}$. The final optimized mask $\textrm{Quant}(\textrm{sigmoid}(\mathbf{m}))$ is employed to remove redundant densified Gaussians and preserve the important ones, which are then
fed into the shift stage.

This selective inheritance mechanism enhances the optimization pipeline in two key aspects. First, the differences between timesteps arise from both temporal transformations and emerging objects. Directly feeding only the base Gaussians into shift stage can lead to inaccurate adjustments in Gaussian properties. By selectively inheriting the previous densified Gaussians before the shift stage, we mitigate these optimization errors.
Second, after the selective inheritance and shift stages, we obtain a refined set of optimized Gaussians, which reduces the number of densified Gaussians requiring optimization, thus improving training efficiency.

\begin{figure}[t]
    \centering
    \includegraphics[width=0.82\linewidth]{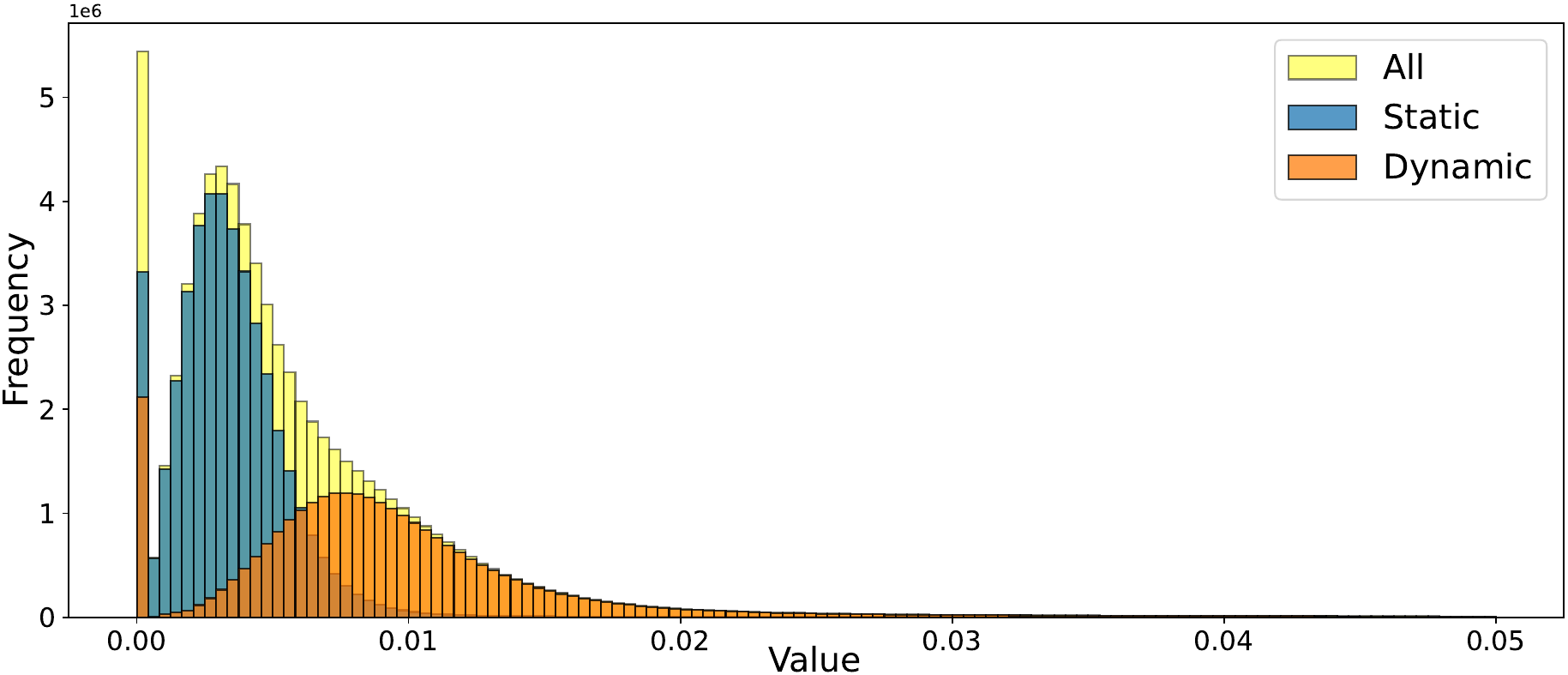}
    \caption{Histogram of Gaussian deformations in the \textit{flame steak} scene of the N3DV dataset. The overall distribution of Gaussian deformations (Yellow) is widely spread, with the majority falling into the low deformation range (less than 0.01). The dynamic (Orange) and static (Blue) components display different deformation patterns, where significant transformations are mainly concentrated in the dynamic component and minimal transformations are primarily found in the static component.}
    \vspace{-4pt}
    \label{fig:flame_steak_hist}
\end{figure}

\subsection{Dynamics-Aware Shift}
\label{sec:shift}
In this stage, we focus on modeling the movements and rotations of Gaussian primitives from the preceding timestep to the current one. A common approach to achieve this is directly learning deformation fields to accommodate temporal transformations, as in previous deformation-based method~\cite{wu20244d} and iterative streaming method~\cite{sun20243dgstream}. However, these approaches often overlook the significant diversity in the movements of Gaussians present in natural scenes, which can result in slower convergence. Specifically, Gaussians representing background or stationary object typically show high similarity with minimal variations or remain unchanged, while those in the foreground, such as humans and other moving objects, display significant dynamics. This disparity is illustrated in Fig. \ref{fig:flame_steak_hist}, presenting a histogram of Gaussian positional offsets between timesteps in the \textit{flame steak} scene of N3DV~\cite{li2022neural}. We observe that most offsets fall within the range of $(0, 0.01]$, which indicates that the majority of Gaussians undergo only minor shifts. Therefore, it is inappropriate to optimize the diverse variations in the scene using a single deformation field.

To address this, we propose estimating a per-Gaussian dynamics mask for all Gaussians before deformation. This mask is used to categorize all primitives into dynamic and static groups. Then, we assign two deformation layers to learn the respective Gaussian transformations: A complex hash-encoding MLP~\cite{muller2022instant} to capture the intricate shifts and rotations in the dynamic group, and a simpler hash-encoding MLP to model the regionally-similar minor variations in the static group.

\begin{figure}[t]
    \centering
    \includegraphics[width=0.85\linewidth]{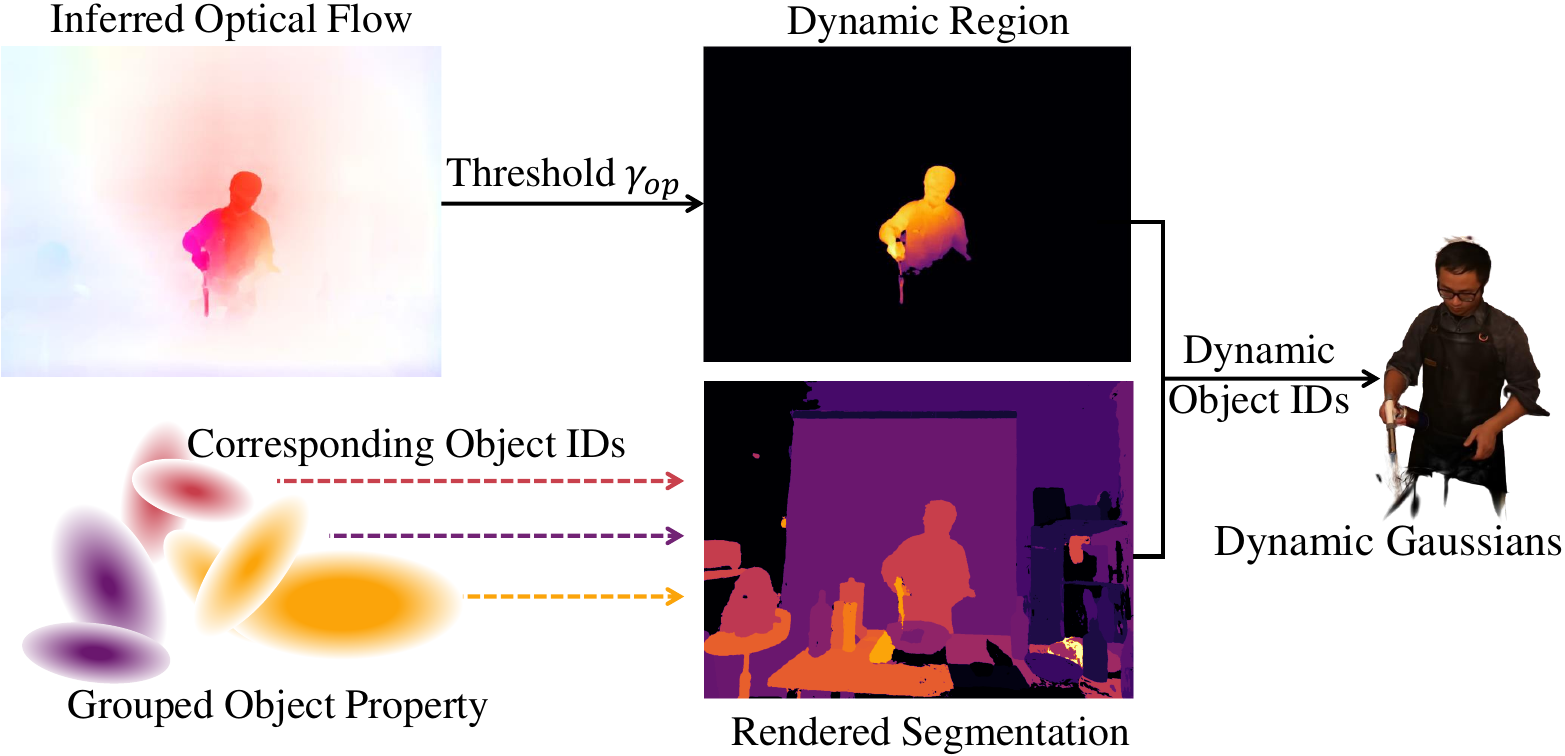}
    \caption{Pipeline of obtaining the per-Gaussian dynamics mask. The process begins with optical flow and segmentation, which provide 2D priors to identify the dynamic object IDs. These IDs are then used to find the corresponding Gaussians in the 3D space.}
    \label{fig:dyna_mask}
    \vspace{-2pt}
\end{figure}

To construct the per-Gaussian dynamics mask before deformation, we leverage techniques from optical flow~\cite{xu2022gmflow,xu2023unifying,wang2025sea} and segmentation~\cite{kirillov2023segment,ke2024segment,ye2025gaussian,shen2025flashsplat}. Specifically, we utilize Gaussian Grouping~\cite{ye2025gaussian} to initialize the scene $\bm{\mathcal{G}}_0$, which assigns an object property, akin to color property, to each Gaussian primitive and optimizes this property with 2D segmentation results as ground truth. These object properties provide per-Gaussian segmentation results that connect 2D images with 3D Gaussians, which facilitates the rendering of segmentation results on any viewpoint. Notably, these object properties remain fixed in subsequent timesteps $\bm{\mathcal{G}}_t (t>0)$, thus posing no additional burden on the time efficiency of on-the-fly training. For subsequent timesteps, we infer an optical flow estimation network~\cite{xu2023unifying} on consecutive timestep frames captured from the same viewpoint and identify the areas where the optical flow exceeds a threshold $\gamma_{op}$, which we classify as dynamic areas. We then query the rendered segmentation results to find out the dynamic object IDs. The dynamic Gaussians are consistently detected using these IDs, as shown in Fig. \ref{fig:dyna_mask}. This process utilizes off-the-shelf methods that do not require network training, allowing us to obtain per-Gaussian dynamics mask within milliseconds (about 300 ms). 

After obtaining the dynamics mask, we employ the multi-resolution hash-encoding layer I-NGP~\cite{muller2022instant} to learn the deformations.
For each Gaussian, the deformation includes a position offset $\bm{\mu}_n\in \mathbb{R}^{3}$, which is added to the Gaussian position $\bm{p}_n$, and a rotation offset $\bm{\sigma}_n\in \mathbb{R}^4$, which is applied to the Gaussian rotation quaternion $\bm{q}_n$ by using $\mathrm{norm}({\bm{q}_n})\times\mathrm{norm}(\bm{\sigma}_n)$. Here, $\mathrm{norm}(\cdot)$ denotes the normalization operation. 
To handle the distinct characteristics of dynamic and static Gaussians, we employ a dual-network strategy: a high-capacity hash-encoding layer $\mathcal{H}_{\textrm{dyn}}$ is used for the dynamic group to capture their complex and varied deformations, while a simpler, lightweight hash-encoding layer $\mathcal{H}_{\textrm{st}}$ is applied to the static group to efficiently manage minor jitters in background and stationary objects. 
This approach not only conserves computational resources but also provides an effective alternative to directly tuning the parameters of all Gaussians. The networks $\mathcal{H}_{\textrm{dyn}}$ and $\mathcal{H}_{\textrm{st}}$ are trained using the fidelity loss function as follows:
\begin{equation}
    \setlength\abovedisplayskip{1pt}
    \setlength\belowdisplayskip{1pt}
    \mathcal{L}=(1-\lambda)\mathcal{L}_1+\lambda\mathcal{L}_{\textrm{D-SSIM}}.
    \label{eq:loss12}
\end{equation}
Fig. \ref{fig:flame_steak_hist} also depicts the optimized deformation distributions of $\mathcal{H}_{\textrm{dyn}}$ and $\mathcal{H}_{\textrm{st}}$ in the \textit{flame steak} scene. The different deformation patterns observed in the dynamic and static groups validate the effectiveness of our proposed strategy.

\begin{figure}[t]
    \centering
    \includegraphics[width=0.92\linewidth]{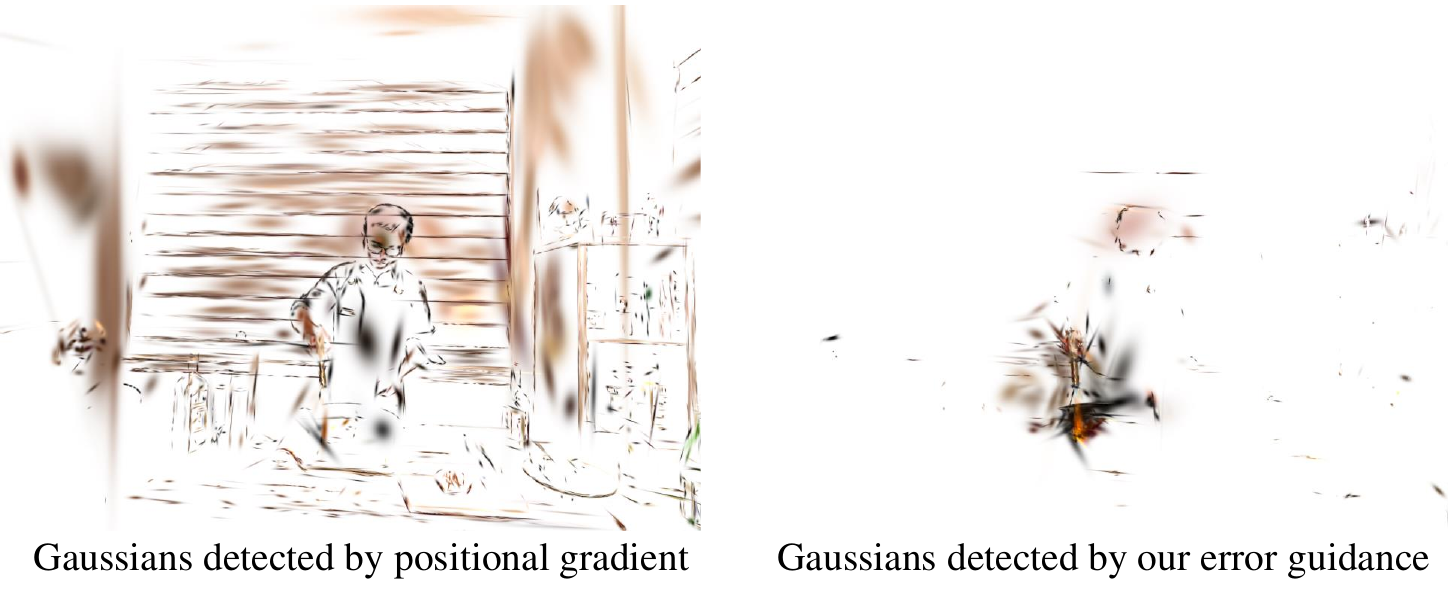}
    \caption{(Left) Gaussians identified for densification based on positional gradients. (Right) Gaussians identified for densification using error-guidance distortion projection. This comparison verifies that our proposed strategy effectively prioritizes emerging objects and achieves targeted compensations. While the vanilla densification strategy requires multiple optimization steps to fully recover the scene, our method concentrates on high-distortion areas with emerging objects, enhancing the computational efficiency.}
    \label{fig:error-guidance}
    \vspace{-4pt}
\end{figure}

\subsection{Error-Guided Densification}
\label{sec:densi}
The above two stages focus on maintaining temporal continuity and managing deformations without introducing new Gaussians. Besides these two stages, densification remains essential to address areas with insufficient reconstruction quality and to incorporate new objects.

To achieve fast and high-quality densification, it is crucial to identify the subset of Gaussians in need of densification. Previous methods~\cite{kerbl20233d,sun20243dgstream} detect under-reconstructed Gaussians based on historical view-space positional gradients, denoted by  $\overline{\nabla \mathbf{p}} \triangleq \{\overline{\nabla p_n}, \forall n\} \in \mathbb{R}^{N\times 1}$. This approach tracks gradients from previous training steps and densifies Gaussians whose positional gradients exceed a specific threshold $\tau_{\textrm{pos}}$.
While this criterion effectively identifies general regions with inadequate reconstruction, as shown in Fig. \ref{fig:error-guidance} (Left), it fails to provide sufficient emphasis on regions with emerging objects or significant reconstruction errors.
This limitation can hinder the efficient optimization of emerging objects, potentially slowing the convergence of on-the-fly training. To address this, we propose an error-adaptive densification strategy that incorporates an additional indicator to identify Gaussians requiring densification. 
This approach projects high-distortion areas from the image space to the 3D space and adaptively enhances densification in critical areas.

Specifically, we collect 2D error maps by comparing the ground truth with the rendered results in the shift stage for each training viewpoint, denoted by $\mathbf{E}^c$, $c=1, 2, \cdots, C$. Then, we filter out the areas with severe distortion above $\gamma_{\textrm{err}}$, yielding an $H\times W$ binary matrix $\mathbf{D}_{\textrm{err}}^c=\{\mathbf{1}[e_i^c>\gamma_{\textrm{err}}],\forall e^c_i\in \mathbf{E}^c, i = 1, \cdots, H \times W \}$ in the 2D pixel plane, where $\mathbf{1}[\cdot]$ is the binary indicator function.
Next, we project the positions of the 3D Gaussians onto the 2D image plane as $(\mathbf{X}, \mathbf{Y}) = \text{Proj}(\bm{\mathcal{G}}^{\textrm{base}})$, where $\mathbf{X} \triangleq \{\mathbf{x}_n, \forall n\}, \mathbf{Y} \triangleq \{\mathbf{y}_n, \forall n\}$, with $(\mathbf{x}_n, \mathbf{y}_n)$ representing the 2D pixel coordinate of the $n$-th Gaussian primitive along the $H$ and $W$ axes. The projection $\text{Proj}(\cdot)$ utilizes the known intrinsic and extrinsic camera matrices, with further details provided in the supplementary material. We then identify the subset of Gaussians that fall within the erroneous areas for each viewpoint, represented as $\mathcal{S}^c_{\textrm{err}}=\{\bm{\mathcal{G}}^{\textrm{base}}|\mathbf{D}_{\textrm{err}}^c(\mathbf{x}_n,\mathbf{y}_n)=1, \forall n\}$.

For these highlighted Gaussian primitives, we apply a relatively lower gradient threshold $\tau_{\textrm{err}}<\tau_{\textrm{pos}}$, thereby placing greater emphasis on high-distortion areas and facilitating more effective error rectification compared to strategies that rely only on positional gradients. This leads to an error-guided adaptive densification scheme, yielding the subset of Gaussians for densification $\mathcal{S}$ as follows:
{\small
\begin{equation}
    \setlength\abovedisplayskip{1pt}
    \mathcal{S}=\{\bm{\mathcal{G}}^{\textrm{base}}|\overline{\nabla p_n}>\tau_{\textrm{pos}},\forall n\} \cup \big\{\mathcal{S}_{\textrm{err}} \cap \{\bm{\mathcal{G}}^{\textrm{base}}|\overline{\nabla p_n}>\tau_{\textrm{err}},\forall n\}\big\},
    \label{eq:error-guidance}
\end{equation}
}where $\mathcal{S}_{\textrm{err}}=\{\cup^C_{c=1}\mathcal{S}^c_{\textrm{err}}\}$ is the combination of high-distortion subsets over all viewpoints (Fig. \ref{fig:error-guidance} (Right)). This error-guided adaptive subset $\mathcal{S}$ prioritizes the defects in historical optimization steps while accommodating under-reconstructed areas. We then perform spawn densification to compensate for both weakly reconstructed regions and emerging objects.
The newly added Gaussians and the transformed inherited Gaussians from $\bm{\mathcal{G}}^{\textrm{var}}_{t-1}$ are optimized using the fidelity loss function in Eq. (\ref{eq:loss12}) to yield $\bm{\mathcal{G}}^{\textrm{var}}_{t}$. During this optimization, the pruning strategy is employed to eliminate excessive Gaussian candidates with very low opacity. Notably, these temporally-varied Gaussians constitute only a small fraction of the total number of primitives in the scene, significantly reducing computational and time costs compared to full-scene optimization.

\begin{table}
    \centering
    \caption{Quantitative comparison on the N3DV dataset.
    The training time and reconstruction qualities are averaged over all 300 frames for each scene.
    $^\dagger$DyNeRF only reports metrics for the \textit{flame salmon} scene.
    $^\ddagger$STG trains each model using a 50-frame video sequence, requiring six models to complete the representation.}
    \resizebox{\columnwidth}{!}{%
    \begin{tabular}{@{}c|l|cccc|c@{}}
      \toprule
      % \hline
      \multirow{2}{*}{Category} & \multirow{2}{*}{Method}                 & PSNR$\uparrow$ & \multirow{2}{*}{DSSIM$\downarrow$} & Train             & Render$\uparrow$       & \multirow{2}{*}{Streamable}\\
                                &                                         & (dB)           &                                    & Time$\downarrow$  & (FPS)                  & \\
      \midrule
      % \hline
      \multirow{2}{*}{Static}   & Plenoxels~\cite{fridovich2022plenoxels} & 30.77          & -            & $>$1000 s     & 8.3          & {\color{red}$\checkmark$}  \\
                                & I-NGP~\cite{muller2022instant}          & 28.62          & -            & 79 s     & 2.9          & {\color{red}$\checkmark$}   \\
      \midrule
      % \hline
      \multirow{9}{*}{Offline}  & DyNeRF$^\dagger$~\cite{li2022neural}    & 29.58            & -          & $>$1000 s     & 0.02         & $\times$ \\
                                & NeRFPlayer~\cite{song2023nerfplayer}    & 30.69            & 0.0340      & 75 s        & 0.05         & {\color{red}$\checkmark$} \\
                                & HexPlane~\cite{cao2023hexplane}         & 31.70            & -          & 140 s       & 0.21         & $\times$ \\
                                & K-Planes~\cite{fridovich2023k}          & 31.63            & -          & 48 s         & 0.15         & $\times$ \\
                                & HyperReel~\cite{attal2023hyperreel}     & 31.10            & 0.0360     & 115 s        & 2         & $\times$ \\
                                & MixVoxels~\cite{wang2023mixed}          & 30.80            & -          & 16 s        & 16.7         & $\times$ \\
                                & 4DGS-Wu~\cite{wu20244d}                 & 31.15            & 0.0331      & \cellcolor{rank3}10.85 s           & 30           & $\times$ \\
                                & 4DGS-Yang~\cite{yangreal}               & \cellcolor{rank2}32.01            & 0.0290      & 76 s           & 114           & $\times$ \\
                                & E-D3DGS~\cite{bae2024per}               & 31.31            & \cellcolor{rank1}0.0259      & 22.4 s           & 74.5         & $\times$ \\
                                & STG$^\ddagger$~\cite{li2024spacetime}              & \cellcolor{rank1}32.05            & \cellcolor{rank2}0.0261      & 60 s           & 140          & $\times$ \\
      \midrule
      % \hline
      \multirow{3}{*}{Online}   & StreamRF~\cite{li2022streaming}         & 30.71            & -          & 14.20 s        & 9 & {\color{red}$\checkmark$}  \\
                                & 3DGStream~\cite{sun20243dgstream}       & 31.68            & 0.0299          & 12.41 s        & \cellcolor{rank2}215 & {\color{red}$\checkmark$}  \\
                                & HiCoM~\cite{gao2025hicom}               & 31.30            & 0.0316          & \cellcolor{rank2}10.22 s & \cellcolor{rank1}230  &  {\color{red}$\checkmark$} \\
                                & DASS (Ours)                                    & \cellcolor{rank3}31.99             & \cellcolor{rank3}0.0285          & \cellcolor{rank1}9.62 s        & \cellcolor{rank2}207   & {\color{red}$\checkmark$}  \\
      \bottomrule
      % \hline
    \end{tabular}
    }
    \label{tab:N3DV_Comparisons_Avg}
    \vspace{-5mm}
\end{table}

\begin{figure*}
    \centering
    \includegraphics[width=0.89\linewidth]{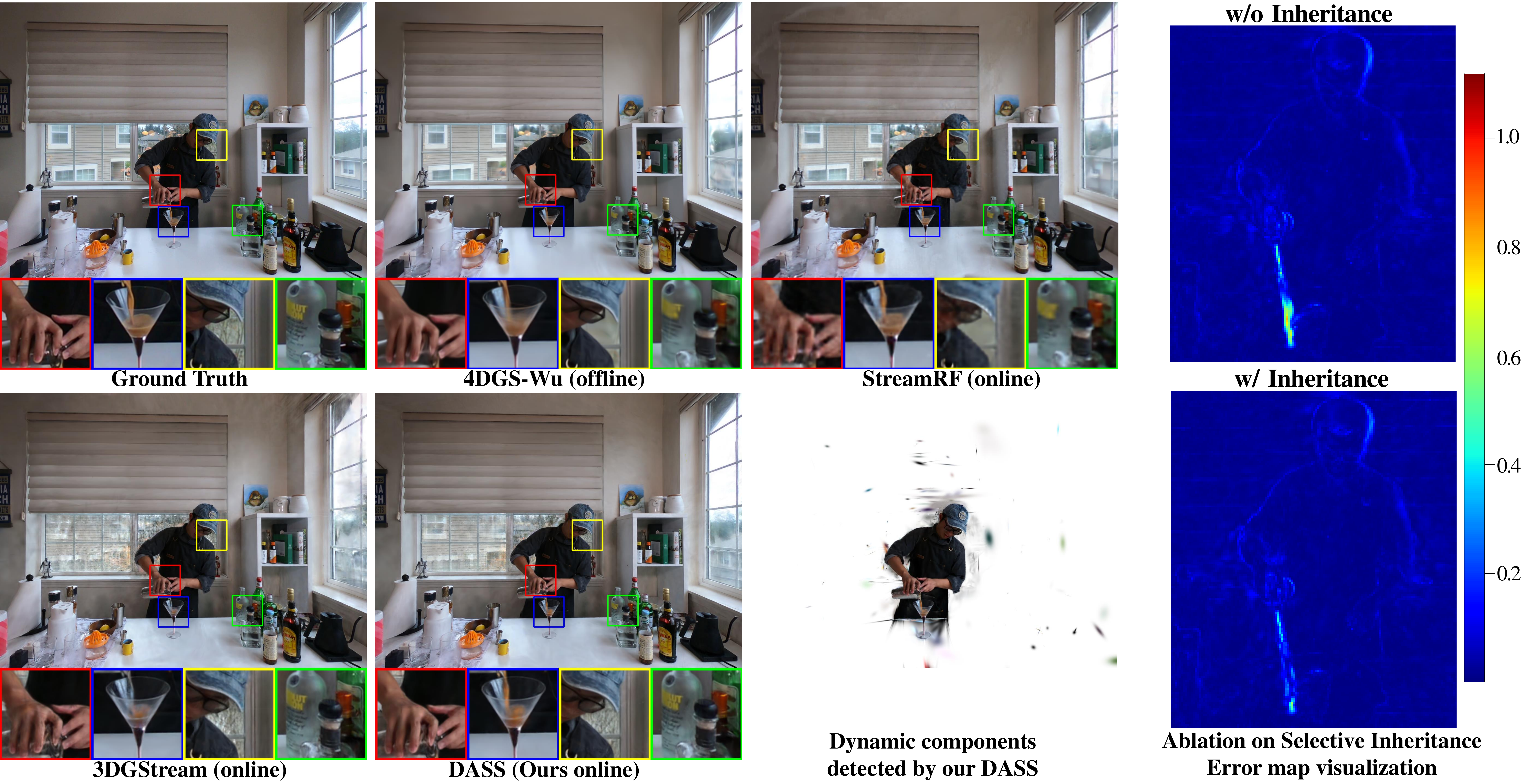}
    \caption{(Left) Qualitative comparison for the scene \textit{coffee martini} of N3DV dataset, including online training methods~\cite{li2022streaming,sun20243dgstream} and offline reconstruction method~\cite{wu20244d}. Compared to other methods, our method accurately recovers dynamic objects and preserves intricate details. (Right) Error map visualization without (top) and with (bottom) the proposed selective inheritance mechanism, on scene \textit{flame steak}. The selective inheritance effectively compensates the artifacts in temporally consistent areas and mitigates the optimization distortions.}
    \label{fig:vis-coffee-n-ablation}
\end{figure*}

\section{Experiments}
\label{sec:exp}
\subsection{Experimental Setup}
\noindent \textbf{Datasets.} To evaluate the effectiveness of our pipeline, we employ two real-world benchmark datasets that are representative of 4D scene reconstruction and streaming tasks. 
(1) \textbf{Neural 3D Video (N3DV)} dataset~\cite{li2022neural} comprises six multi-view videos captured from 18 to 21 viewpoints, each at a resolution of $2704\times 2028$. Following previous methods~\cite{li2022neural,li2022streaming,sun20243dgstream,yangreal,zhang2024mega}, our experiments are conducted at a half resolution of $1352\times 1014$, using one view for evaluation and the remaining views for training. 
(2) \textbf{Meet Room} dataset, provided by the previous work on the streaming task~\cite{li2022streaming}, includes videos at a resolution of $1280\times 720$ and a frame rate of 30 FPS. This dataset is captured from 13 viewpoint cameras, which is relatively sparser than N3DV. Consistent with the baselines~\cite{li2022streaming,sun20243dgstream}, one view is reserved for testing and the other twelve views are used for training.

\noindent \textbf{Implementation.} 
We initiate our training at timestep 0 using Gaussian Grouping~\cite{ye2025gaussian} and optimize the object properties for segmentation concurrently during this initialization, after which these properties remain fixed for subsequent timesteps. The optical flow is directly inferred using Unimatch~\cite{xu2023unifying}. These off-the-shelf methods obtain the segmentation and optical flow results without extra pre-training, thereby imposing minimal computational or temporal overhead during on-the-fly training and streaming. For subsequent timesteps, the optimization processes for the inheritance, shift, and densification stages are executed with $s_1=20$, $s_2=100$, and $s_3=60$ steps, respectively.

\noindent \textbf{Baselines.} Our primary comparisons are against iterative 4D dynamic spatial reconstruction methods that support on-the-fly training and per-timestep streaming, specifically StreamRF~\cite{li2022streaming},  3DGStream~\cite{sun20243dgstream} and HiCoM~\cite{gao2025hicom}, which are categorized as online methods. To ensure fair comparisons, we evaluate these online methods on the same RTX A6000 GPU.
To evaluate reconstruction fidelity, we also include other 4D dynamic spatial reconstruction methods that require full-length multi-view videos as input and do not support on-the-fly training or per-timestep streaming. This offline category includes NeRF-based methods such as DyNeRF~\cite{li2022neural}, NeRFPlayer~\cite{song2023nerfplayer}, HexPlane~\cite{cao2023hexplane}, K-Planes~\cite{fridovich2023k}, HyperReel~\cite{attal2023hyperreel}, and MixVoxels~\cite{wang2023mixed}, as well as recent 3DGS-based methods like 4DGS-Wu~\cite{wu20244d}, 4DGS-Yang~\cite{yangreal}, E-D3DGS~\cite{bae2024per}, and STG~\cite{li2024spacetime}. Note that NeRFPlayer~\cite{song2023nerfplayer} allows streaming after optimization but does not support on-the-fly training.
In addition, we consider static methods, including Plenoxels~\cite{2plenoxels} and I-NGP~\cite{muller2022instant}, which fully train a static scene representation for each timestep.

\noindent \textbf{Metrics.} We focus on fast on-the-fly training for streamable dynamic spatial reconstruction, emphasizing both time efficiency and reconstruction fidelity. To evaluate time efficiency, we calculate the average per-timestep training time across the entire video sequence and measure the frames per second (FPS). For offline methods that do not support on-the-fly training, we report their training time averaged over all timesteps. To assess reconstruction fidelity, we use metrics including peak signal-to-noise ratio (PSNR) and dissimilarity structural similarity index measure (DSSIM). 

\subsection{Results and Comparisons}
The quantitative evaluation results of our method on the N3DV dataset~\cite{li2022neural} are detailed in Tab. \ref{tab:N3DV_Comparisons_Avg}, and a comprehensive comparison of multiple metrics is shown in Fig. \ref{fig:teaser} (Middle). Our method achieves the fastest on-the-fly reconstruction speed, which converges from the previous timestep to the current one within 10 seconds, yielding a significant improvement in efficiency. In contrast, the average training time of baseline methods ranges from tens of seconds to several minutes. 
This significant time efficiency advantage originates from several key designs: 
First, our inheritance stage effectively captures the temporal continuity and selectively preserves the optimized results from the previous timestep. This significantly alleviates the optimization burden in subsequent stages, especially in the densification stage, which requires only 60 steps compared to the 100 steps in 3DGStream. Moreover, the selective inheritance stage is highly lightweight, as it optimizes only one learnable parameter for each Gaussian primitive, in contrast to the tens of parameters required in a fully tuned pipeline.
In addition, our dynamics-aware shift stage assigns different deformation fields for dynamic and static components, focusing on significant and subtle variations, respectively. This strategy enhances the convergence and  achieves high-quality deformations with fewer optimization steps (100 steps instead of 150 steps in 3DGStream). 
Although our pipeline requires optical flow and segmentation, these results are either directly inferred or rendered within milliseconds and are executed only once every few timesteps. 

Beyond the significant efficiency advantage, our reconstruction fidelity also outperforms previous online streamable baselines~\cite{li2022streaming,sun20243dgstream,gao2025hicom} and surpasses most state-of-the-art offline training methods~\cite{li2022neural,song2023nerfplayer,cao2023hexplane,fridovich2023k,attal2023hyperreel,wang2023mixed,wu20244d,bae2024per}.
Our dynamics-aware deformation places greater emphasis on areas with complex motions and variations, thereby facilitating more detailed reconstruction. Besides, our error-guided densification strategy identifies areas with weak construction and newly emerging objects, which rapidly detects and mitigates the reconstruction defects. 
Qualitative comparisons on the N3DV dataset, along with a visualization of our detected dynamics components, are provided in Fig. \ref{fig:vis-coffee-n-ablation} (Left). A detailed comparison of each scene on the N3DV dataset is provided in the supplementary material.

Our rendering speed further highlights the superiority of our method over others. Leveraging the real-time 3DGS rasterization, our rendering speed significantly outperforms that of NeRF-based methods. Moreover, our method achieves higher FPS than 4DGS methods, since our deformation fields, comprising only hash encoding and simple MLPs, enable a low query overhead of around 1 ms.

Similarly, our method demonstrates fast on-the-fly training, efficient rendering speed, and high-fidelity reconstruction on the MeetRoom dataset, as detailed in Tab. \ref{tab:meetroom_Comparisons}.

\begin{table}
    \centering
    \caption{Quantitative comparison on the Meet Room dataset, with metrics averaged over all frames. Following previous baselines, the scene \textit{Discussion} is chosen for evaluation. 
    }
    
    % \resizebox{\columnwidth}{!}{%
    {\scriptsize
    \begin{tabular}{@{}l|ccc@{}}
    \toprule
    Method  & PSNR (dB)$\uparrow$& Train Time (s)$\downarrow$   & Render (FPS)$\uparrow$\\
    \midrule
    Plenoxels~\cite{fridovich2022plenoxels} & 27.15 & 840   & 10     \\
    I-NGP~\cite{muller2022instant}      & 28.10 & 65    & 4.1   \\
    \hline
    StreamRF~\cite{li2022streaming}  & 27.03 & 9.97    & 12     \\
    3DGStream~\cite{sun20243dgstream} & \underline{30.79} & 6.20 & \textbf{275}    \\
    HiCoM~\cite{gao2025hicom} & 29.87 & \underline{5.58} & \textbf{277} \\
    DASS (Ours)                              & \textbf{31.02}   & \textbf{4.91}      & \underline{254} \\
    \bottomrule
    \end{tabular}}
    % }
    \label{tab:meetroom_Comparisons}
    \vspace{-5mm}
\end{table}

\subsection{Ablation Studies}
We conduct ablation experiments on the N3DV dataset to evaluate the contributions of each component in our proposed design. We start from a baseline optimizing Gaussian positions and progressively transition to our full model, with all variants maintaining consistent optimization steps. The results regarding training time and reconstruction fidelity are summarized in Tab. \ref{tab:ablation}, which demonstrates the effectiveness of our designs in improving reconstruction quality while minimizing training time. Additional ablation studies are provided in Supplementary Sec. 8.1.

\noindent \textbf{Effectiveness of Dynamics-Aware Shift.} To validate the improvement brought by our dynamics-aware shift strategy, we compare two variants: one with a uniform shift deformation field (+ Shift) and another with dynamics-aware deformation (+ Dynamics aware), as referenced in Tab. \ref{tab:ablation}. The latter achieves a better PSNR performance with comparable time consumption, which underscores the importance of our dynamics-aware optimization in distinguishing between intrinsic dynamic and static components and effectively managing the complex movements.

\noindent \textbf{Effectiveness of Error-Guided Densification.} 
To evaluate the effectiveness of our error-guided densification, we consider three variants in Tab. \ref{tab:ablation}. A plain densification stage, similar to pervious works~\cite{kerbl20233d,sun20243dgstream} (+ Densification), yields modest performance improvements. Simply lowering the densification threshold $\tau_{\textrm{pos}}$ (+ Lower threshold) results in an excessive number of Gaussians, increasing training time by 0.79s. In contrast, our proposed error-guided densification strategy (+ Error guidance), which integrates $\mathcal{S}_{\textrm{err}}$ as in Eq. \ref{eq:error-guidance}, adaptively focuses on regions with weak reconstruction and emerging objects, providing superior performance.

\noindent \textbf{Effectiveness of Selective Inheritance.} 
To evaluate the impact of the selective inheritance stage, we compare two variants in Tab. \ref{tab:ablation}. Directly inheriting all Gaussians from previous timestep (+ Direct inheritance) leads to excessive accumulation of Gaussians, which hinders fast on-the-fly optimization and introduces errors to the reconstruction. Conversely, our selective inheritance stage (+ Selective inheritance) effectively exploits the temporal consistency, balancing the preservation of essential Gaussians while controlling numerical accumulation. We also visualize the error maps before and after the selective inheritance stage in Fig. \ref{fig:vis-coffee-n-ablation} (Right), which demonstrates that our selective inheritance strategy effectively recovers temporal coherence with highly lightweight optimization. By optimizing just one parameter per Gaussian for $s_3=20$ steps, our approach reduces the optimization overhead for subsequent stages.

\begin{table}[t]
    \centering
    \caption{Results of ablation experiments, which are averaged over six scenes from the N3DV dataset.}
    % \vspace{-11 pt}
    \resizebox{\linewidth}{!}{
    \begin{tabular}{ccc|ccc}
        \toprule
        Method & PSNR (dB)$\uparrow$ & Time (s)$\downarrow$ & Method & PSNR (dB)$\uparrow$ & Time (s)$\downarrow$ \\ \midrule
        Baseline & 30.60 & 11.35 & + Lower threshold & 31.68 & 10.78\\
        + Shift & 30.95 & 9.97 & + Error guidance & 31.76 & 10.03 \\ 
        + Dynamics aware & 31.31 & 10.08 & + Direct inheritance & 31.71 & 11.88 \\ 
        + Densification & 31.60 & 9.99 & + Selective inheritance & 31.99 & 9.62 \\ \bottomrule
    \end{tabular}
    }
    \label{tab:ablation}
    \vspace{-15pt}
\end{table}

\section{Conclusion}
\label{sec:conclusion}
In this paper, we introduce a novel three-stage pipeline for iterative online 4D dynamic spatial reconstruction, which allows for on-the-fly training and per-timestep streaming. We incorporate a selective inheritance stage to capture the temporal continuity, a dynamics-aware shift stage to exploit the dynamic and static features in natural scenes, and an error-guided densification stage to adaptively recover emerging objects and weak reconstruction. Our method achieves state-of-the-art performance in online streaming 4D reconstruction, providing the fastest training speed and superior representation quality.

{
    \small
    \bibliographystyle{ieeenat_fullname}
    \bibliography{main}

\begin{thebibliography}{57}
\providecommand{\natexlab}[1]{#1}
\providecommand{\url}[1]{\texttt{#1}}
\expandafter\ifx\csname urlstyle\endcsname\relax
  \providecommand{\doi}[1]{doi: #1}\else
  \providecommand{\doi}{doi: \begingroup \urlstyle{rm}\Url}\fi

\bibitem[Arena et~al.(2022)Arena, Collotta, Pau, and Termine]{arena2022overview}
Fabio Arena, Mario Collotta, Giovanni Pau, and Francesco Termine.
\newblock An overview of augmented reality.
\newblock \emph{Computers}, 11\penalty0 (2):\penalty0 28, 2022.

\bibitem[Attal et~al.(2023)Attal, Huang, Richardt, Zollhoefer, Kopf, O’Toole, and Kim]{attal2023hyperreel}
Benjamin Attal, Jia-Bin Huang, Christian Richardt, Michael Zollhoefer, Johannes Kopf, Matthew O’Toole, and Changil Kim.
\newblock Hyperreel: High-fidelity 6-dof video with ray-conditioned sampling.
\newblock In \emph{Proceedings of the IEEE/CVF Conference on Computer Vision and Pattern Recognition}, pages 16610--16620, 2023.

\bibitem[Bae et~al.(2024)Bae, Kim, Yun, Lee, Bang, and Uh]{bae2024per}
Jeongmin Bae, Seoha Kim, Youngsik Yun, Hahyun Lee, Gun Bang, and Youngjung Uh.
\newblock Per-{G}aussian embedding-based deformation for deformable 3{D} {G}aussian splatting.
\newblock In \emph{European Conference on Computer Vision}, 2024.

\bibitem[Bengio et~al.(2013)Bengio, L{\'e}onard, and Courville]{bengio2013estimating}
Yoshua Bengio, Nicholas L{\'e}onard, and Aaron Courville.
\newblock Estimating or propagating gradients through stochastic neurons for conditional computation.
\newblock \emph{arXiv preprint arXiv:1308.3432}, 2013.

\bibitem[Cao and Johnson(2023)]{cao2023hexplane}
Ang Cao and Justin Johnson.
\newblock Hexplane: A fast representation for dynamic scenes.
\newblock In \emph{Proceedings of the IEEE/CVF Conference on Computer Vision and Pattern Recognition}, pages 130--141, 2023.

\bibitem[Chen et~al.(2024{\natexlab{a}})Chen, Chen, Zhang, Wang, Yang, Wang, Cai, Yang, Liu, and Lin]{chen2024gaussianeditor}
Yiwen Chen, Zilong Chen, Chi Zhang, Feng Wang, Xiaofeng Yang, Yikai Wang, Zhongang Cai, Lei Yang, Huaping Liu, and Guosheng Lin.
\newblock {G}aussianeditor: Swift and controllable 3{d} editing with {G}aussian splatting.
\newblock In \emph{Proceedings of the IEEE/CVF Conference on Computer Vision and Pattern Recognition}, pages 21476--21485, 2024{\natexlab{a}}.

\bibitem[Chen et~al.(2024{\natexlab{b}})Chen, Wu, Cai, Harandi, and Lin]{chen2024hac}
Yihang Chen, Qianyi Wu, Jianfei Cai, Mehrtash Harandi, and Weiyao Lin.
\newblock {HAC}: Hash-grid assisted context for 3{D} {G}aussian splatting compression.
\newblock In \emph{European Conference on Computer Vision}, 2024{\natexlab{b}}.

\bibitem[Cheng et~al.(2023)Cheng, Oh, Price, Schwing, and Lee]{cheng2023tracking}
Ho~Kei Cheng, Seoung~Wug Oh, Brian Price, Alexander Schwing, and Joon-Young Lee.
\newblock Tracking anything with decoupled video segmentation.
\newblock In \emph{Proceedings of the IEEE/CVF International Conference on Computer Vision}, pages 1316--1326, 2023.

\bibitem[Duan et~al.(2024)Duan, Wei, Dai, He, Chen, and Chen]{duan20244d}
Yuanxing Duan, Fangyin Wei, Qiyu Dai, Yuhang He, Wenzheng Chen, and Baoquan Chen.
\newblock 4d-rotor gaussian splatting: towards efficient novel view synthesis for dynamic scenes.
\newblock In \emph{ACM SIGGRAPH 2024 Conference Papers}, pages 1--11, 2024.

\bibitem[Fan et~al.(2023)Fan, Wang, Wen, Zhu, Xu, and Wang]{fan2023lightgaussian}
Zhiwen Fan, Kevin Wang, Kairun Wen, Zehao Zhu, Dejia Xu, and Zhangyang Wang.
\newblock Light{G}aussian: Unbounded 3{D} {G}aussian compression with 15x reduction and 200+ fps.
\newblock \emph{arXiv preprint arXiv:2311.17245}, 2023.

\bibitem[Fan et~al.(2024)Fan, Cong, Wen, Wang, Zhang, Ding, Xu, Ivanovic, Pavone, Pavlakos, et~al.]{fan2024instantsplat}
Zhiwen Fan, Wenyan Cong, Kairun Wen, Kevin Wang, Jian Zhang, Xinghao Ding, Danfei Xu, Boris Ivanovic, Marco Pavone, Georgios Pavlakos, et~al.
\newblock Instantsplat: Unbounded sparse-view pose-free gaussian splatting in 40 seconds.
\newblock \emph{arXiv preprint arXiv:2403.20309}, 2\penalty0 (3):\penalty0 4, 2024.

\bibitem[Fridovich-Keil et~al.(2022{\natexlab{a}})Fridovich-Keil, Yu, Tancik, Chen, Recht, and Kanazawa]{2plenoxels}
Sara Fridovich-Keil, Alex Yu, Matthew Tancik, Qinhong Chen, Benjamin Recht, and Angjoo Kanazawa.
\newblock Plenoxels: Radiance fields without neural networks.
\newblock In \emph{Proceedings of the IEEE/CVF Conference on Computer Vision and Pattern Recognition}, pages 5501--5510, 2022{\natexlab{a}}.

\bibitem[Fridovich-Keil et~al.(2022{\natexlab{b}})Fridovich-Keil, Yu, Tancik, Chen, Recht, and Kanazawa]{fridovich2022plenoxels}
Sara Fridovich-Keil, Alex Yu, Matthew Tancik, Qinhong Chen, Benjamin Recht, and Angjoo Kanazawa.
\newblock Plenoxels: Radiance fields without neural networks.
\newblock In \emph{Proceedings of the IEEE/CVF conference on computer vision and pattern recognition}, pages 5501--5510, 2022{\natexlab{b}}.

\bibitem[Fridovich-Keil et~al.(2023)Fridovich-Keil, Meanti, Warburg, Recht, and Kanazawa]{fridovich2023k}
Sara Fridovich-Keil, Giacomo Meanti, Frederik~Rahb{\ae}k Warburg, Benjamin Recht, and Angjoo Kanazawa.
\newblock K-planes: Explicit radiance fields in space, time, and appearance.
\newblock In \emph{Proceedings of the IEEE/CVF Conference on Computer Vision and Pattern Recognition}, pages 12479--12488, 2023.

\bibitem[Gao et~al.(2025)Gao, Meng, Wen, Chen, and Zhang]{gao2025hicom}
Qiankun Gao, Jiarui Meng, Chengxiang Wen, Jie Chen, and Jian Zhang.
\newblock Hicom: Hierarchical coherent motion for dynamic streamable scenes with 3d gaussian splatting.
\newblock \emph{Advances in Neural Information Processing Systems}, 37:\penalty0 80609--80633, 2025.

\bibitem[Guo et~al.(2024)Guo, Zhou, Li, Wang, and Li]{guo2024motion}
Zhiyang Guo, Wengang Zhou, Li Li, Min Wang, and Houqiang Li.
\newblock Motion-aware 3d {G}aussian splatting for efficient dynamic scene reconstruction.
\newblock \emph{arXiv preprint arXiv:2403.11447}, 2024.

\bibitem[He et~al.(2024)He, Chen, Lu, Song, and Zhang]{he2024s4d}
Bing He, Yunuo Chen, Guo Lu, Li Song, and Wenjun Zhang.
\newblock S4{D}: Streaming 4{D} real-world reconstruction with {G}aussians and 3d control points.
\newblock \emph{arXiv preprint arXiv:2408.13036}, 2024.

\bibitem[Kajiya and Von~Herzen(1984)]{kajiya1984ray}
James~T Kajiya and Brian~P Von~Herzen.
\newblock Ray tracing volume densities.
\newblock \emph{ACM SIGGRAPH computer graphics}, 18\penalty0 (3):\penalty0 165--174, 1984.

\bibitem[Katsumata et~al.(2024)Katsumata, Vo, and Nakayama]{katsumata2024compact}
Kai Katsumata, Duc~Minh Vo, and Hideki Nakayama.
\newblock A compact dynamic 3{D} {G}aussian representation for real-time dynamic view synthesis.
\newblock In \emph{European Conference on Computer Vision}, 2024.

\bibitem[Ke et~al.(2024)Ke, Ye, Danelljan, Tai, Tang, Yu, et~al.]{ke2024segment}
Lei Ke, Mingqiao Ye, Martin Danelljan, Yu-Wing Tai, Chi-Keung Tang, Fisher Yu, et~al.
\newblock Segment anything in high quality.
\newblock \emph{Advances in Neural Information Processing Systems}, 36, 2024.

\bibitem[Kerbl et~al.(2023)Kerbl, Kopanas, Leimkuehler, and Drettakis]{kerbl20233d}
Bernhard Kerbl, Georgios Kopanas, Thomas Leimkuehler, and George Drettakis.
\newblock 3{D} {G}aussian splatting for real-time radiance field rendering.
\newblock \emph{ACM Transactions on Graphics (TOG)}, 42\penalty0 (4):\penalty0 1--14, 2023.

\bibitem[Kheradmand et~al.(2024)Kheradmand, Rebain, Sharma, Sun, Tseng, Isack, Kar, Tagliasacchi, and Yi]{kheradmand20243d}
Shakiba Kheradmand, Daniel Rebain, Gopal Sharma, Weiwei Sun, Jeff Tseng, Hossam Isack, Abhishek Kar, Andrea Tagliasacchi, and Kwang~Moo Yi.
\newblock 3{D} {G}aussian splatting as markov chain monte carlo.
\newblock \emph{arXiv preprint arXiv:2404.09591}, 2024.

\bibitem[Kirillov et~al.(2023)Kirillov, Mintun, Ravi, Mao, Rolland, Gustafson, Xiao, Whitehead, Berg, Lo, et~al.]{kirillov2023segment}
Alexander Kirillov, Eric Mintun, Nikhila Ravi, Hanzi Mao, Chloe Rolland, Laura Gustafson, Tete Xiao, Spencer Whitehead, Alexander~C Berg, Wan-Yen Lo, et~al.
\newblock Segment anything.
\newblock In \emph{Proceedings of the IEEE/CVF International Conference on Computer Vision}, pages 4015--4026, 2023.

\bibitem[Lee et~al.(2024)Lee, Rho, Sun, Ko, and Park]{lee2024compact}
Joo~Chan Lee, Daniel Rho, Xiangyu Sun, Jong~Hwan Ko, and Eunbyung Park.
\newblock Compact 3{D} {G}aussian representation for radiance field.
\newblock In \emph{Proceedings of the IEEE/CVF Conference on Computer Vision and Pattern Recognition}, pages 21719--21728, 2024.

\bibitem[Li et~al.(2022{\natexlab{a}})Li, Shen, Wang, Shen, and Tan]{li2022streaming}
Lingzhi Li, Zhen Shen, Zhongshu Wang, Li Shen, and Ping Tan.
\newblock Streaming radiance fields for 3{D} video synthesis.
\newblock \emph{Advances in Neural Information Processing Systems}, 35:\penalty0 13485--13498, 2022{\natexlab{a}}.

\bibitem[Li et~al.(2022{\natexlab{b}})Li, Slavcheva, Zollhoefer, Green, Lassner, Kim, Schmidt, Lovegrove, Goesele, Newcombe, et~al.]{li2022neural}
Tianye Li, Mira Slavcheva, Michael Zollhoefer, Simon Green, Christoph Lassner, Changil Kim, Tanner Schmidt, Steven Lovegrove, Michael Goesele, Richard Newcombe, et~al.
\newblock Neural 3{D} video synthesis from multi-view video.
\newblock In \emph{Proceedings of the IEEE/CVF Conference on Computer Vision and Pattern Recognition}, pages 5521--5531, 2022{\natexlab{b}}.

\bibitem[Li et~al.(2024)Li, Chen, Li, and Xu]{li2024spacetime}
Zhan Li, Zhang Chen, Zhong Li, and Yi Xu.
\newblock Spacetime {G}aussian feature splatting for real-time dynamic view synthesis.
\newblock In \emph{Proceedings of the IEEE/CVF Conference on Computer Vision and Pattern Recognition}, pages 8508--8520, 2024.

\bibitem[Liu and Banerjee(2024)]{liu2024swings}
Bangya Liu and Suman Banerjee.
\newblock Swings: Sliding window {G}aussian splatting for volumetric video streaming with arbitrary length.
\newblock \emph{arXiv preprint arXiv:2409.07759}, 2024.

\bibitem[Liu et~al.(2024)Liu, Wu, Zhang, Wang, Li, and Kwong]{liu2024compgs}
Xiangrui Liu, Xinju Wu, Pingping Zhang, Shiqi Wang, Zhu Li, and Sam Kwong.
\newblock Comp{GS}: Efficient 3{D} scene representation via compressed {G}aussian splatting.
\newblock \emph{arXiv preprint arXiv:2404.09458}, 2024.

\bibitem[Lu et~al.(2024{\natexlab{a}})Lu, Yu, Xu, Xiangli, Wang, Lin, and Dai]{lu2024scaffold}
Tao Lu, Mulin Yu, Linning Xu, Yuanbo Xiangli, Limin Wang, Dahua Lin, and Bo Dai.
\newblock Scaffold-{GS}: Structured 3{D} {G}aussians for view-adaptive rendering.
\newblock In \emph{Proceedings of the IEEE/CVF Conference on Computer Vision and Pattern Recognition}, pages 20654--20664, 2024{\natexlab{a}}.

\bibitem[Lu et~al.(2024{\natexlab{b}})Lu, Guo, Hui, Chen, Yang, Tang, Zhu, and Dai]{lu20243d}
Zhicheng Lu, Xiang Guo, Le Hui, Tianrui Chen, Min Yang, Xiao Tang, Feng Zhu, and Yuchao Dai.
\newblock 3{D} geometry-aware deformable {G}aussian splatting for dynamic view synthesis.
\newblock In \emph{Proceedings of the IEEE/CVF Conference on Computer Vision and Pattern Recognition}, pages 8900--8910, 2024{\natexlab{b}}.

\bibitem[Mildenhall et~al.(2021)Mildenhall, Srinivasan, Tancik, Barron, Ramamoorthi, and Ng]{nerf}
Ben Mildenhall, Pratul~P Srinivasan, Matthew Tancik, Jonathan~T Barron, Ravi Ramamoorthi, and Ren Ng.
\newblock {N}e{RF}: Representing scenes as neural radiance fields for view synthesis.
\newblock \emph{Communications of the ACM}, 65\penalty0 (1):\penalty0 99--106, 2021.

\bibitem[M{\"u}ller et~al.(2022)M{\"u}ller, Evans, Schied, and Keller]{muller2022instant}
Thomas M{\"u}ller, Alex Evans, Christoph Schied, and Alexander Keller.
\newblock Instant neural graphics primitives with a multiresolution hash encoding.
\newblock \emph{ACM transactions on graphics (TOG)}, 41\penalty0 (4):\penalty0 1--15, 2022.

\bibitem[Pumarola et~al.(2021)Pumarola, Corona, Pons-Moll, and Moreno-Noguer]{DNERF}
Albert Pumarola, Enric Corona, Gerard Pons-Moll, and Francesc Moreno-Noguer.
\newblock D-{N}e{RF}: Neural radiance fields for dynamic scenes.
\newblock In \emph{Proceedings of the IEEE/CVF Conference on Computer Vision and Pattern Recognition}, pages 10318--10327, 2021.

\bibitem[Shen et~al.(2025)Shen, Yang, and Wang]{shen2025flashsplat}
Qiuhong Shen, Xingyi Yang, and Xinchao Wang.
\newblock Flashsplat: 2{D} to 3{D} {G}aussian splatting segmentation solved optimally.
\newblock In \emph{European Conference on Computer Vision}, pages 456--472. Springer, 2025.

\bibitem[Song et~al.(2023)Song, Chen, Li, Chen, Chen, Yuan, Xu, and Geiger]{song2023nerfplayer}
Liangchen Song, Anpei Chen, Zhong Li, Zhang Chen, Lele Chen, Junsong Yuan, Yi Xu, and Andreas Geiger.
\newblock {N}e{RF}player: A streamable dynamic scene representation with decomposed neural radiance fields.
\newblock \emph{IEEE Transactions on Visualization and Computer Graphics}, 29\penalty0 (5):\penalty0 2732--2742, 2023.

\bibitem[Sun et~al.(2024{\natexlab{a}})Sun, Jiao, Li, Zhang, Zhao, and Xing]{sun20243dgstream}
Jiakai Sun, Han Jiao, Guangyuan Li, Zhanjie Zhang, Lei Zhao, and Wei Xing.
\newblock 3{DGS}tream: On-the-fly training of 3{D} {G}aussians for efficient streaming of photo-realistic free-viewpoint videos.
\newblock In \emph{Proceedings of the IEEE/CVF Conference on Computer Vision and Pattern Recognition}, pages 20675--20685, 2024{\natexlab{a}}.

\bibitem[Sun et~al.(2024{\natexlab{b}})Sun, Shi, Ooi, Huang, and Hsu]{sun2024multi}
Yuan-Chun Sun, Yuang Shi, Wei~Tsang Ooi, Chun-Ying Huang, and Cheng-Hsin Hsu.
\newblock Multi-frame bitrate allocation of dynamic 3{D} {G}aussian splatting streaming over dynamic networks.
\newblock In \emph{Proceedings of the 2024 SIGCOMM Workshop on Emerging Multimedia Systems}, pages 1--7, 2024{\natexlab{b}}.

\bibitem[Teed and Deng(2020)]{teed2020raft}
Zachary Teed and Jia Deng.
\newblock Raft: Recurrent all-pairs field transforms for optical flow.
\newblock In \emph{Computer Vision--ECCV 2020: 16th European Conference, Glasgow, UK, August 23--28, 2020, Proceedings, Part II 16}, pages 402--419. Springer, 2020.

\bibitem[Tu et~al.(2024)Tu, Shao, Dong, Zheng, Zhang, Chen, Wang, Li, Ma, Zhang, et~al.]{tu2024tele}
Hanzhang Tu, Ruizhi Shao, Xue Dong, Shunyuan Zheng, Hao Zhang, Lili Chen, Meili Wang, Wenyu Li, Siyan Ma, Shengping Zhang, et~al.
\newblock Tele-{A}loha: A telepresence system with low-budget and high-authenticity using sparse rgb cameras.
\newblock In \emph{ACM SIGGRAPH 2024 Conference Papers}, pages 1--12, 2024.

\bibitem[Wang et~al.(2023)Wang, Tan, Li, Tian, Song, and Liu]{wang2023mixed}
Feng Wang, Sinan Tan, Xinghang Li, Zeyue Tian, Yafei Song, and Huaping Liu.
\newblock Mixed neural voxels for fast multi-view video synthesis.
\newblock In \emph{Proceedings of the IEEE/CVF International Conference on Computer Vision}, pages 19706--19716, 2023.

\bibitem[Wang et~al.(2024{\natexlab{a}})Wang, Zhang, Wang, Yao, Xie, Yu, Wu, and Xu]{wang2024v}
Penghao Wang, Zhirui Zhang, Liao Wang, Kaixin Yao, Siyuan Xie, Jingyi Yu, Minye Wu, and Lan Xu.
\newblock V\^{} 3: Viewing volumetric videos on mobiles via streamable 2d dynamic {G}aussians.
\newblock \emph{arXiv preprint arXiv:2409.13648}, 2024{\natexlab{a}}.

\bibitem[Wang et~al.(2024{\natexlab{b}})Wang, Wu, Zhang, and Xu]{wang2024gscream}
Yuxin Wang, Qianyi Wu, Guofeng Zhang, and Dan Xu.
\newblock {GS}cream: Learning 3{D} geometry and feature consistent {G}aussian splatting for object removal.
\newblock \emph{arXiv preprint arXiv:2404.13679}, 2024{\natexlab{b}}.

\bibitem[Wang et~al.(2025)Wang, Lipson, and Deng]{wang2025sea}
Yihan Wang, Lahav Lipson, and Jia Deng.
\newblock Sea-raft: Simple, efficient, accurate raft for optical flow.
\newblock In \emph{European Conference on Computer Vision}, pages 36--54. Springer, 2025.

\bibitem[Wang et~al.(2003)Wang, Simoncelli, and Bovik]{wang2003multiscale}
Zhou Wang, Eero~P Simoncelli, and Alan~C Bovik.
\newblock Multiscale structural similarity for image quality assessment.
\newblock In \emph{The Thrity-Seventh Asilomar Conference on Signals, Systems \& Computers, 2003}, pages 1398--1402. Ieee, 2003.

\bibitem[Wu et~al.(2024)Wu, Yi, Fang, Xie, Zhang, Wei, Liu, Tian, and Wang]{wu20244d}
Guanjun Wu, Taoran Yi, Jiemin Fang, Lingxi Xie, Xiaopeng Zhang, Wei Wei, Wenyu Liu, Qi Tian, and Xinggang Wang.
\newblock 4{D} {G}aussian splatting for real-time dynamic scene rendering.
\newblock In \emph{Proceedings of the IEEE/CVF Conference on Computer Vision and Pattern Recognition}, pages 20310--20320, 2024.

\bibitem[Xu et~al.(2022{\natexlab{a}})Xu, Zhang, Cai, Rezatofighi, and Tao]{xu2022gmflow}
Haofei Xu, Jing Zhang, Jianfei Cai, Hamid Rezatofighi, and Dacheng Tao.
\newblock Gmflow: Learning optical flow via global matching.
\newblock In \emph{Proceedings of the IEEE/CVF conference on computer vision and pattern recognition}, pages 8121--8130, 2022{\natexlab{a}}.

\bibitem[Xu et~al.(2023)Xu, Zhang, Cai, Rezatofighi, Yu, Tao, and Geiger]{xu2023unifying}
Haofei Xu, Jing Zhang, Jianfei Cai, Hamid Rezatofighi, Fisher Yu, Dacheng Tao, and Andreas Geiger.
\newblock Unifying flow, stereo and depth estimation.
\newblock \emph{IEEE Transactions on Pattern Analysis and Machine Intelligence}, 2023.

\bibitem[Xu et~al.(2022{\natexlab{b}})Xu, Xu, Philip, Bi, Shu, Sunkavalli, and Neumann]{xu2022point}
Qiangeng Xu, Zexiang Xu, Julien Philip, Sai Bi, Zhixin Shu, Kalyan Sunkavalli, and Ulrich Neumann.
\newblock Point-{N}e{RF}: Point-based neural radiance fields.
\newblock In \emph{Proceedings of the IEEE/CVF conference on computer vision and pattern recognition}, pages 5438--5448, 2022{\natexlab{b}}.

\bibitem[Yang et~al.(2024{\natexlab{a}})Yang, Gao, Zhou, Jiao, Zhang, and Jin]{yang2024deformable}
Ziyi Yang, Xinyu Gao, Wen Zhou, Shaohui Jiao, Yuqing Zhang, and Xiaogang Jin.
\newblock Deformable 3{D} {G}aussians for high-fidelity monocular dynamic scene reconstruction.
\newblock In \emph{Proceedings of the IEEE/CVF Conference on Computer Vision and Pattern Recognition}, pages 20331--20341, 2024{\natexlab{a}}.

\bibitem[Yang et~al.(2024{\natexlab{b}})Yang, Yang, Pan, and Zhang]{yangreal}
Zeyu Yang, Hongye Yang, Zijie Pan, and Li Zhang.
\newblock Real-time photorealistic dynamic scene representation and rendering with 4{D} {G}aussian splatting.
\newblock In \emph{The Twelfth International Conference on Learning Representations}, 2024{\natexlab{b}}.

\bibitem[Ye et~al.(2025{\natexlab{a}})Ye, Danelljan, Yu, and Ke]{ye2025gaussian}
Mingqiao Ye, Martin Danelljan, Fisher Yu, and Lei Ke.
\newblock Gaussian grouping: Segment and edit anything in 3d scenes.
\newblock In \emph{European Conference on Computer Vision}, pages 162--179. Springer, 2025{\natexlab{a}}.

\bibitem[Ye et~al.(2025{\natexlab{b}})Ye, Wan, Li, Hong, Li, Li, Zhang, and Lin]{ye20253d}
Zhifan Ye, Chenxi Wan, Chaojian Li, Jihoon Hong, Sixu Li, Leshu Li, Yongan Zhang, and Yingyan~Celine Lin.
\newblock 3{D} {G}aussian rendering can be sparser: Efficient rendering via learned fragment pruning.
\newblock \emph{Advances in Neural Information Processing Systems}, 37:\penalty0 5850--5869, 2025{\natexlab{b}}.

\bibitem[Zhang et~al.(2024)Zhang, Liu, Zhang, Ge, He, Xu, Wang, Lin, Yan, and Zhang]{zhang2024mega}
Xinjie Zhang, Zhening Liu, Yifan Zhang, Xingtong Ge, Dailan He, Tongda Xu, Yan Wang, Zehong Lin, Shuicheng Yan, and Jun Zhang.
\newblock Mega: Memory-efficient 4{D} {G}aussian splatting for dynamic scenes.
\newblock \emph{arXiv preprint arXiv:2410.13613}, 2024.

\bibitem[Zhang et~al.(2025)Zhang, Ge, Xu, He, Wang, Qin, Lu, Geng, and Zhang]{zhang2025gaussianimage}
Xinjie Zhang, Xingtong Ge, Tongda Xu, Dailan He, Yan Wang, Hongwei Qin, Guo Lu, Jing Geng, and Jun Zhang.
\newblock Gaussianimage: 1000 fps image representation and compression by 2{D} {G}aussian splatting.
\newblock In \emph{European Conference on Computer Vision}, pages 327--345. Springer, 2025.

\bibitem[Zhou et~al.(2024)Zhou, Chang, Jiang, Fan, Zhu, Xu, Chari, You, Wang, and Kadambi]{zhou2024feature}
Shijie Zhou, Haoran Chang, Sicheng Jiang, Zhiwen Fan, Zehao Zhu, Dejia Xu, Pradyumna Chari, Suya You, Zhangyang Wang, and Achuta Kadambi.
\newblock Feature 3dgs: Supercharging 3d gaussian splatting to enable distilled feature fields.
\newblock In \emph{Proceedings of the IEEE/CVF Conference on Computer Vision and Pattern Recognition}, pages 21676--21685, 2024.

\bibitem[Zwicker et~al.(2001)Zwicker, Pfister, Van~Baar, and Gross]{zwicker2001ewa}
Matthias Zwicker, Hanspeter Pfister, Jeroen Van~Baar, and Markus Gross.
\newblock {EWA} volume splatting.
\newblock In \emph{Proceedings Visualization, 2001. VIS'01.}, pages 29--538. IEEE, 2001.

\end{thebibliography}
}

\newpage
\section*{Appendix}
\appendix

The supplementary material provides an introduction of preliminary knowledge in Sec. \ref{sec:Prelimi}, more implementation details in
Sec. \ref{sec:imp}, additional experimental results in Sec. \ref{sec:EXP}, analysis on concurrent works in Sec. \ref{sec:concurrent}, and discusses potential extensions in Sec. \ref{sec:discuss}.

\section{Preliminaries}
\label{sec:Prelimi}
\subsection{3D Gaussian Splatting}
3D Gaussian Splatting (3DGS)~\cite{kerbl20233d} exploits a point-based representation to reconstruct 3D space from multi-view image inputs. A Gaussian primitive is expressed as:
\begin{equation} 
    G(\bm{x}) = e^{-\frac{1}{2}(\bm{x}-\bm{\mu})^T \bm{\Sigma}^{-1} (\bm{x}-\bm{\mu})}, 
    \label{3dgs}
\end{equation}
where $\bm{\mu} \in \mathbb{R}^3$ and $\bm{\Sigma} \in \mathbb{R}^{3 \times 3}$ denote the position vector and the covariance matrix, respectively. The covariance matrix $\bm{\Sigma}$ is decomposed into a rotation matrix $\mathbf{R} \in \mathbb{R}^{3\times3}$ and a scaling matrix $\mathbf{S} \in \mathbb{R}^{3\times3}$ as:
\begin{equation}
    \bm{\Sigma} = \mathbf{R}\mathbf{S}\mathbf{S}^T\mathbf{R}^T,
\end{equation}
where $\mathbf{R}$ and $\mathbf{S}$ are further represented by a quaternion $\mathbf{q} \in \mathbb{R}^{4}$ and a vector $\mathbf{s} \in \mathbb{R}^{3}$, respectively. When rendering an image on a 2D plane, this covariance matrix from the 3D space is projected onto the 2D plane as follows:
\begin{equation}
    \bm{\Sigma}_{\textrm{2D}} = \mathbf{J}\mathbf{W}\bm{\Sigma}\mathbf{W}^T\mathbf{J}^T,
\end{equation}
where $\mathbf{W}$ is the viewing projection matrix and $\mathbf{J}$ is the Jacobian of the affine approximation for the projective transformation. After reconstructing a 3D space with $N$ Gaussian primitives, the rendered RGB image, denoted by $\bm{C}$, is obtained by blending the contributing Gaussians in depth order as follows:
\begin{equation}
    \bm{C}=\sum_{i=1}^N\bm{c}_i\alpha_i\prod_{j=1}^{i-1}(1-\alpha_j),
\end{equation}
where $\alpha_i$ denotes the product of the 2D Gaussian with the $i$-th greatest depth and its opacity property $o_i$, and $\bm{c}_i$ represents its view-dependent color property.

\subsection{Gaussian Grouping}
As elaborated in the previous subsection, the vanilla 3DGS method assigns position, scale, rotation, opacity, and color attributes to each Gaussian primitive. In addition to these attributes, Gaussian Grouping~\cite{ye2025gaussian} introduces an additional set of learnable properties $\bm{e}_i \in \mathbb{R}^{16}$ to each Gaussian for segmentation, which is referred to as Identity Encoding or object property. Similar to the color property, this object property facilitates rendering on the 2D image plane as follows:
\begin{equation}
    \mathbf{M}=\sum_{i=1}^N\bm{e}_i\alpha_i\prod_{j=1}^{i-1}(1-\alpha_j),
\end{equation}
where $\mathbf{M}$ represents the rendered 2D mask identity feature, which is of the same size as the rendered RGB image. To optimize the object property $\bm{e}_i$, Gaussian Grouping first utilizes 2D segmentation method~\cite{kirillov2023segment} to generate segmentation masks for each multi-view image, and then employs a well-trained zero-shot tracker~\cite{cheng2023tracking} to maintain consistent segmentation masks across frames. These 2D segmentation masks supervise the optimization of the object property $\bm{e}_i$ and a single convolution layer to recover the object property dimension into segmentation classes. This optimization is performed concurrently with the optimization of other visual properties. In this way, each Gaussian primitive is assigned an object property that groups the Gaussians into segmented objects. In our proposed DASS, this Gaussian Grouping technique serves as a bridge between the 2D and 3D representations, which facilitates the identification of dynamic object IDs in the 2D image and their corresponding dynamic 3D Gaussian primitives.

\subsection{Optical Flow}
Optical flow is a representation of the apparent relative motion between consecutive video frames and used to describe the dynamics and variations on the 2D image plane. Among the abundant advanced methods, we employ the Transformer-based optical flow estimation method~\cite{xu2023unifying}. This approach balances accuracy and efficiency without the need of iterative refinements, making it suitable for our on-the-fly training setup. 

Specifically, the optical flow estimation can be modeled as finding 2D pixel-wise correspondences between consecutive images on a 2D plane. The images are downsampled and enhanced in the feature space, yielding $\mathbf{F}_1,\mathbf{F}_2\in \mathbb{R}^{\frac{H}{8}\times\frac{W}{8}\times D}$. Then, the feature similarity is calculated as follows:
\begin{equation}
    \mathbf{S}=\frac{\mathbf{F}_1\mathbf{F}_2^T}{\sqrt{D}}.
\end{equation}
Next, a softmax function is applied to $\mathbf{S}$, yielding a normalized matching distribution for each position in $\mathbf{F}_1$ corresponding to $\mathbf{F}_2$, represented as $\textrm{softmax}(\mathbf{S})$. The correspondence relationship is then derived by multiplying the pixel grid $\mathbf{U}_{\mathrm{2D}}$ with the matching distribution as: 
\begin{equation}
    \hat{\mathbf{U}}_{\mathrm{2D}}=\textrm{softmax}(\mathbf{S})\cdot \mathbf{U}_\mathrm{2D}.
\end{equation}
This leads to the optical flow estimation by calculating the discrepancy between the corresponding pixel coordinates as:
\begin{equation}
    \mathbf{V}=\hat{\mathbf{U}}_{\mathrm{2D}}-\mathbf{U}_\mathrm{2D}.
\end{equation}
An additional self-attention layer is used to propagate and remedy unreliable results in this estimation. To get the original image resolution prediction, RAFT’s upsampling~\cite{teed2020raft} method is utilized to compute the full-resolution optical flow.

\section{Additional Implementation Details}
\label{sec:imp}
\subsection{Implementation and Metrics}
We implement our streaming pipeline based on the official open-source codebase of 3DGStream~\cite{sun20243dgstream} 
and strictly follow the evaluation principles. Specifically, the reported training time \textbf{includes} the time consumption for \textbf{training the initialization at frame 0}. This timestep 0 initialization process takes 10.18 minutes in our pipeline and is counted into the average training time calculation following the official setup of 3DGStream. Notably, the time consumption of subsequent on-the-fly training takes only \textbf{7.59 s} for each timestep. The experiments are conducted on an RTX A6000 GPU and an Intel Xeon Platinum 8370C CPU.

\subsection{Additional Details in Each Stage}
\noindent \textbf{Initialization.} For the initialization at timestep 0, we employ the official codebase of Gaussian Grouping. The segmentation-related hyperparameters in Gaussian Grouping remain unmodified, as we find that varying these hyperparameters leads to minimal impact on the streaming performance.  
Moreover, during training the initialization, we introduce an opacity regularization term, written as $-\frac{1}{N}\sum^N_{i=1}o_i\log(o_i)$, in addition to the fidelity loss. This opacity regularization encourages the opacity of Gaussian primitives to approach either zero or one, thereby pushing Gaussians to the object surface and naturally pruning nearly transparent Gaussians. 
This regularization has proven effective in previous works~\cite{duan20244d,zhang2024mega} to enhance the representation efficiency and quality.

\noindent \textbf{Selective Inheritance.} In our proposed selective inheritance stage, the quantized learnable vector after the sigmoid function, $\textrm{Quant}(\textrm{sigmoid}(\mathbf{m}))$, determines whether to maintain the densified Gaussian primitives from the previous timestep. Notably, the quantization operation itself is non-differentiable. Therefore, the Straight Through Estimator (STE)~\cite{bengio2013estimating} is applied in the optimization of $\mathbf{m}$, which is written as: 
{\small
\begin{equation}
    \mathbf{m}_{\textrm{op}}=\textrm{detach}(\textrm{Quant}(\textrm{sigmoid}(\mathbf{m}))-\textrm{sigmoid}(\mathbf{m}))+\textrm{sigmoid}(\mathbf{m}),
\end{equation}
}where $\mathbf{m}_{\textrm{op}}$ is the parameter involved in differentiable optimization. Moreover, this selective inheritance is paused every 20 timesteps, i.e., the inherited Gaussians are erased once for every 20 timesteps. This strategy considers the typical duration of temporally consistent objects and avoids introducing misleading information during optimization. 

\noindent \textbf{Dynamics-Aware Shift.} The dynamics-aware shift stage distinguishes the dynamic and static groups of Gaussian primitives in the representation and employs different hash-encoding MLPs, $\mathcal{H}_{\textrm{dyn}}$ and $\mathcal{H}_{\textrm{st}}$, as deformation fields for their optimization. 
The estimation of dynamic objects usually remains consistent across timesteps, so repeatedly performing this estimation introduces temporal redundancy into the on-the-fly training. In our pipeline, the estimation of dynamic Gaussian group (as shown in Fig. 4 in the main body of the paper) is performed every 10 timesteps, considering that the dynamic properties remain stable across adjacent timesteps. 
As discussed in Sec. 3.3, the expressive ability of I-NGP is affected by the network complexity parameters, including the hash table size $T_{\textrm{Hash}}$ and the number of feature dimensions per entry $F_{\textrm{Hash}}$. 
Subsequently, the hash table size $T_{\textrm{Hash}}$ and the number of feature dimensions per entry $F_{\textrm{Hash}}$ vary between dynamic and static groups. Specifically, for the N3DV dataset, the dynamic group adopts $T_{\textrm{Hash}}=2^{16}$ and $F_{\textrm{Hash}}=4$ for $\mathcal{H}_{\textrm{dyn}}$, while the static group uses $T_{\textrm{Hash}}=2^{14}$ and $F_{\textrm{Hash}}=2$ for $\mathcal{H}_{\textrm{st}}$. For the Meet Room dataset, the dynamic group employs $T_{\textrm{Hash}}=2^{15}$ and $F_{\textrm{Hash}}=4$, while the static group adopts $T_{\textrm{Hash}}=2^{13}$ and $F_{\textrm{Hash}}=2$.

\noindent \textbf{Error-Guided Densification.} In the error-guided densification stage, we identify the Gaussians whose positions fall into high-distortion areas for densification to improve reconstruction and recover emerging objects. The projection of Gaussian positions is detailed in Algorithm \ref{alg:projection}.

\begin{algorithm}[h]
    \renewcommand{\algorithmicrequire}{\textbf{Input:}}
    \renewcommand{\algorithmicensure}{\textbf{Output:}}
    \caption{Projection of 3D Gaussian positions to 2D image plane pixels}
    \label{alg:projection}
    \begin{algorithmic}[1]
        \REQUIRE Positions of $N$ 3D Gaussians $\mathbf{P}=[\mathbf{X},\mathbf{Y},\mathbf{Z}]\in \mathbb{R}^{N\times 3}$, rendered image size $H$ and $W$, full projection matrix $\mathbf{T}\in \mathbb{R}^{4\times 4}$ embedded in the 3DGS camera model, which is composed of the camera intrinsic matrix and the extrinsic matrix.
        \STATE Calculate the projection in homography space as $\mathbf{P}_{\textrm{hom}}=[\mathbf{P},\mathbf{1}]\cdot \mathbf{T}=[\mathbf{p}_{x},\mathbf{p}_{y},\mathbf{p}_{z},\mathbf{p}_{\textrm{hom}}]$.
        \STATE Normalize the coordinates on the image plane as $\mathbf{x}_{\textrm{norm}}=\mathbf{p}_{x}/\mathbf{p}_{\textrm{hom}}$ and $\mathbf{y}_{\textrm{norm}}=\mathbf{p}_{y}/\mathbf{p}_{\textrm{hom}}$
        \STATE Obtain the corresponding pixel-level coordinates as $\mathbf{x}_n=\textrm{Round}(0.5\cdot((\mathbf{x}_{\textrm{norm}}+1)\cdot$W$-1))$ and $\mathbf{y}_n=\textrm{Round}(0.5\cdot((\mathbf{x}_{\textrm{norm}}+1)\cdot $H$-1))$
        \ENSURE The corresponding pixels of the 3D Gaussian positions as $(\mathbf{X},\mathbf{Y})=[\mathbf{x}_n,\mathbf{y}_n]$.
    \end{algorithmic}
\end{algorithm}

\section{Additional Experimental Results}
\label{sec:EXP}
\subsection{Additional Ablation Studies}
\noindent \textbf{Impact of Optimization Steps.}
To assess the impact of the optimization steps at each stage and illustrate the convergence efficiency of our DASS framework, we adjust the optimization steps and evaluate the reconstruction qualities, as shown in Tab. \ref{tab:opt-steps}. Our DASS assigns $s_1=20$, $s_2=100$, and $s_3=60$ for the inheritance, shift, and densification stages, respectively, serving as the baseline in the first row of Tab. \ref{tab:opt-steps}. 
This assignment contributes to the time efficiency of our method compared to the state-of-the-art baseline 3DGStream~\cite{sun20243dgstream} (9.62 s compared to 12 s), which assigns a total of 250 optimization steps (150 for transformation and 100 for densification). The baselines with increased optimization steps shown in subsequent rows verify that our DASS achieves optimal convergence with the primary optimization setting. Increasing the optimization steps leads to a negligible improvement in representation quality, less than 0.05 dB, while significantly increasing time consumption, thereby validating our convergence efficiency. This efficiency comes from multiple aspects: 
Our selective inheritance stage preserves the essential Gaussian primitives of the previous timestep's densified Gaussians and mitigates the optimization distortions with only $s_3=20$ steps of optimization. The dynamics-aware shift stage learns the diverse distributions of dynamic and static components, respectively, facilitating the deformation convergence. Besides, assigning a low-complexity deformation field to the static components conserves temporal and computational resources for the majority of the Gaussian primitives. The error-guided densification stage effectively detects and compensates for the distortions, enhancing the reconstruction quality while introducing low temporal overhead.
In addition, the last two baselines in Tab. \ref{tab:opt-steps} demonstrate that reducing the training steps leads to inferior reconstruction quality, necessitating our optimization step assignment.

\begin{table}[htbp]
  \centering
  \caption{Ablation study on the numbers of optimization steps, with the results averaged over all six scenes of the N3DV dataset~\cite{li2022neural}.}
    \resizebox{\linewidth}{!}{
    \begin{tabular}{ccccc}
    \toprule
    Inheritance & Shift & Densification & PSNR$\uparrow$ & Time$\downarrow$ \\
    s1 & s2 & s3 & (dB) & (s) \\
    \midrule
    20    & 100   & 60    & 31.99 & 9.62 \\
    \midrule
    50    & 100   & 60    & 31.96 & 10.29 \\
    20    & 150   & 60    & 32.01 & 11.78 \\
    20    & 100   & 100   & 31.99 & 10.79 \\
    20    & 150   & 100   & 32.03 & 12.53 \\
    20    & 50    & 60    & 31.26 & 7.41 \\
    20    & 100   & 30    & 31.50  & 8.73 \\
    \bottomrule
    \end{tabular}}
  \label{tab:opt-steps}
\end{table}

\begin{table}[t]
  \centering
  \caption{Comparison with naive 3DGS methods. The training time and reconstruction qualities are averaged over all 300 frames for each scene of the N3DV dataset~\cite{li2022neural}.}
    \begin{tabular}{ccc}
    \toprule
    Baseline & PSNR (dB)$\uparrow$ & Time (s)$\downarrow$ \\
    \midrule
    Naive Per-Timestep 3DGS & 32.07    & 615 \\
    Intuitive Full-Tuned 3DGS & 30.83     & 11.86 \\
    DASS (Ours) & 31.99    & 9.62 \\
    \bottomrule
    \end{tabular}%
    % \vspace{-15pt}
  \label{tab:vanilla3DGS}%
\end{table}%

\subsection{Quantitative Results}
\begin{figure}
    \centering
    \includegraphics[width=0.9\linewidth]{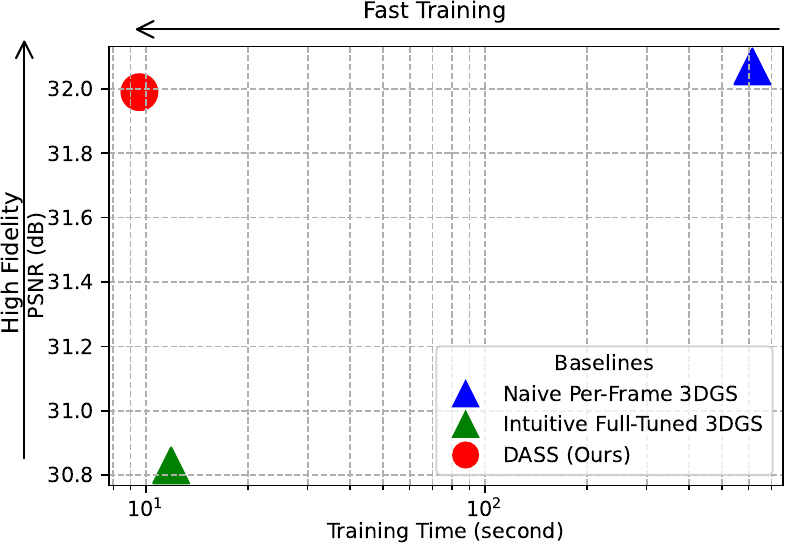}
    \caption{Performance comparisons with two variants of naive 3DGS methods in terms of per-frame training time and reconstruction quality (PSNR) on the N3DV dataset~\cite{li2022neural}. Our method achieves a much better balance between training efficiency and representation quality.}
    % \vspace{-15pt}
    \label{fig:supp-time-psnr}
\end{figure}

In addition to the state-of-the-art 4D reconstruction baselines shown in Tab. 1 in the main body of the paper, we provide comparisons with two variants of naive 3DGS methods in Tab. \ref{tab:vanilla3DGS} and Fig. \ref{fig:supp-time-psnr}. 
Specifically, we compare our method with a naive streamable baseline, which trains an independent set of 3D Gaussian primitives for each timestep, referred to as ``Naive Per-Timestep 3DGS'' in Tab. \ref{tab:vanilla3DGS}. In this baseline, representation of each timestep is trained with the same conditions as the initial zero-timestep in streaming methods. Although this naive baseline outperforms all off-the-shelf baselines in 4D reconstruction, it fully trains a set of Gaussian primitives from scratch for each scene, leading to a significant training time overhead of approximately 615 seconds per-timestep. In contrast, our DASS effectively models the temporal variations in the 3D space and facilitates a fast convergence between timesteps. DASS demonstrates competitive streaming quality, with only a negligible gap of 0.08 dB compared to this naive baseline, while consuming merely 1.56\% of the training time. This significant improvement highlights the efficiency of DASS.
We also compare our method with an intuitive full-tuned baseline, which iteratively tunes all parameters of the Gaussian primitives initialized from the previous timestep, termed ``Intuitive Full-Tuned 3DGS'' in Tab. \ref{tab:vanilla3DGS}. Specifically, it initializes the on-the-fly training with the same timestep 0 initialization as our DASS and tunes all Gaussian properties for a fixed number of iterations, which yields a similar time consumption to ours. As shown in Tab. \ref{tab:vanilla3DGS}, a significant performance degradation of 1.16 dB is observed in this baseline. This indicates that the intuitive baseline fails to capture temporal variations between timesteps under such time efficiency, while our DASS effectively manages and converges these variations with lower temporal overhead.

Besides, we provide the quantitative comparison for each scene of the N3DV dataset~\cite{li2022neural} detailed in Tab. \ref{tab:N3DV_Comparisons}, which is compared with offline training NeRF-based methods~\cite{li2022neural,song2023nerfplayer,cao2023hexplane,fridovich2023k,attal2023hyperreel,wang2023mixed}, offline training 3DGS-based methods~\cite{wu20244d,yangreal,bae2024per,li2024spacetime}, online training NeRF-based method~\cite{li2022streaming}, and online training 3DGS-based method~\cite{sun20243dgstream}.

\subsection{Time Consumption}
The average time consumption for each stage in our DASS framework is provided as follows: Inheritance (0.62 s), Shift (4.02 s), and Densification (1.75 s), with an additional 3.23s spent on fixed expenses, including $\bm{\mathcal{G}}_{0}$ initialization, data loading, and saving. With low training time overhead, our designs efficiently facilitate the transformation of existing elements and retain emerging objects.

\subsection{Qualitative Results}
\begin{figure*}[t]
    \centering
    \includegraphics[width=0.8\linewidth]{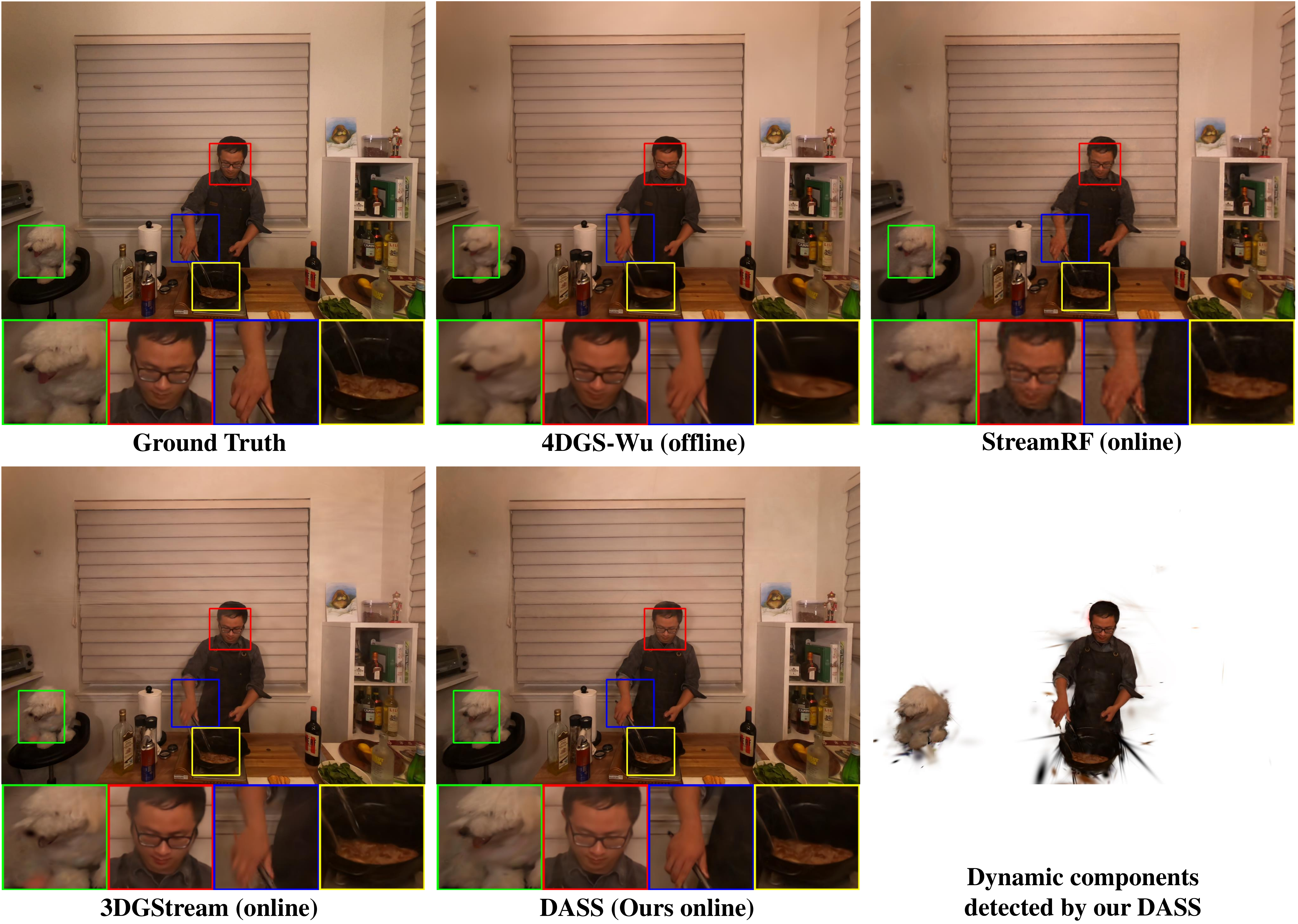}
    % \vspace{-8pt}
    \caption{Qualitative results for the \textit{sear steak} scene. Our method effectively captures the dramatic motions and color changes in the scene. However, the details in the furry dog part (green) and the steak part (yellow) are blurry in 4DGS-Wu, while the color of tongue (green) and the shape of hand (blue) are incorrect in 3DGStream.}
    \label{fig:sear-steak-vis}
\end{figure*}

\begin{figure*}[h]
    \centering
    \includegraphics[width=0.8\linewidth]{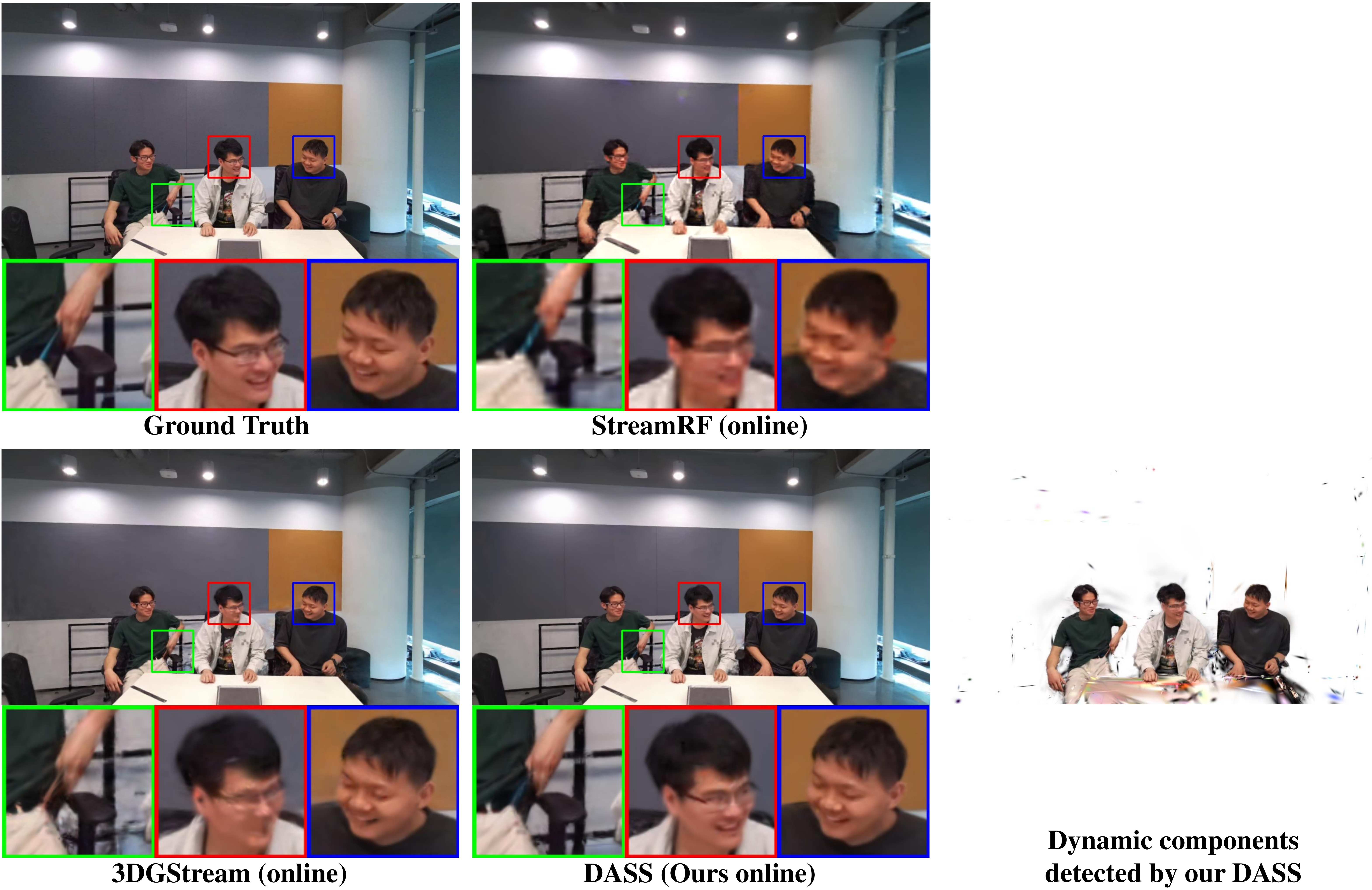}
    % \vspace{-8pt}
    \caption{Qualitative results for the \textit{discussion} scene. The baseline methods experience blurs and errors in dramatically dynamic areas, as seen in the hand moving out of the pocket (green) and in the human faces (red and blue). In contrast, our method preserves better dynamic features and motion representations. We implement 4DGS on the MeetRoom dataset but it fails to yield reasonable reconstruction, probably due to the sparse view challenges in this dataset.}
    \label{fig:discussion-vis}
\end{figure*}

\begin{figure*}[h]
    \centering
    \includegraphics[width=0.8\linewidth]{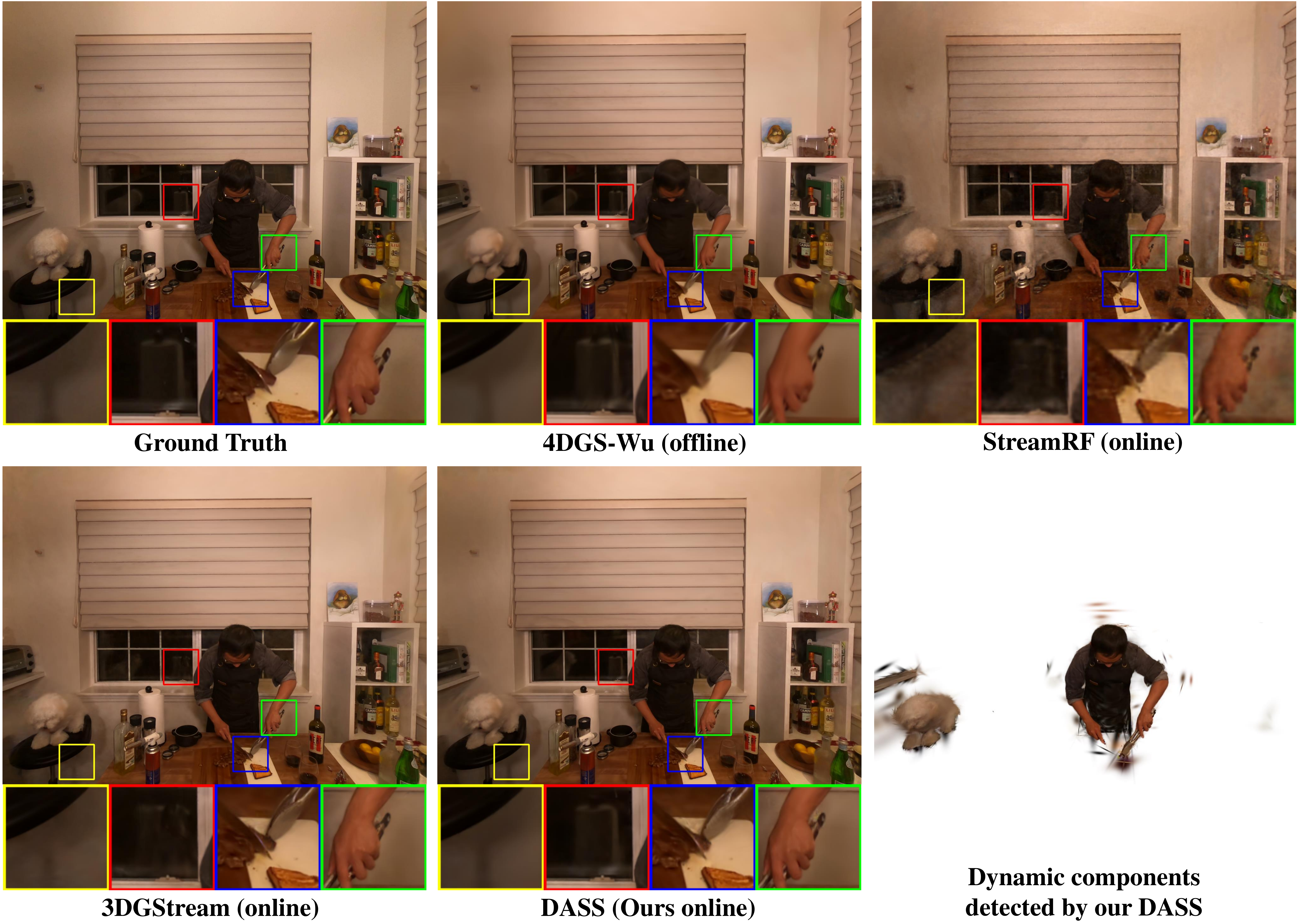}
    % \vspace{-10pt}
    \caption{Qualitative results for the \textit{cut beef} scene. The baselines 4DGS-Wu and StreamRF present over-smooth and lack some details (blue and green), while our DASS faithfully reconstructs finer details. Compared to 3DGStream, our DASS provides more accurate and vivid representations, such as the moving shadow (yellow), the reflection in the window (red), and the color of the knife (blue).}
    \label{fig:cut-beef-vis}
\end{figure*}

We provide additional qualitative results of our method in Fig. \ref{fig:sear-steak-vis}, Fig. \ref{fig:discussion-vis}, and Fig. \ref{fig:cut-beef-vis}. Each figure includes the rendered test view visualizations and comparisons with baseline methods. Our DASS presents high-quality reconstruction and fine visual details compared with the baseline methods. We also provide visualizations of the detected dynamics components, which effectively guides the optimization to focus on dynamic objects and complex motions.

\section{Analysis on Concurrent Works}
\label{sec:concurrent}
We have noticed several concurrent studies on iterative 4D reconstruction, although their focuses and contributions differ from ours. 
For instance, S4D~\cite{he2024s4d} directly manipulates 3D control points to guide the movements, but it requires full optimization that incurs significant time costs on key frames. V$^3$~\cite{wang2024v} aims to facilitate streaming free-viewpoint videos on mobile devices, with a particular emphasis on the reduction of the per-timestep streaming storage overhead. Conversely, our work identifies the per-timestep convergence speed as the critical bottleneck. Moreover, while V$^3$ decomposes motion and appearance attributes for efficient reconstruction, it primarily applies to scenarios with moving objects against vacant background and does not accommodate emerging objects. In addition, SwinGS~\cite{liu2024swings} concentrates on maintaining a consistently stable stream data volume over time and reducing storage overhead through the use of Markov Chain Monte Carlo~\cite{kheradmand20243d} and a window-based design. Another work~\cite{sun2024multi} considers the bit allocation in a system-level implementation. In contrast, our work prioritizes on-the-fly per-timestep training efficiency as the key bottleneck and focuses on obtaining high-quality streaming while minimizing optimization time.

\section{Discussion}
\label{sec:discuss}
Our DASS pipeline effectively addresses the challenge of on-the-fly 4D reconstruction for per-frame streaming by integrating selective inheritance, dynamics-aware shift, and error-guided densification stages. These designs have the potential to broaden applications in various 3DGS-based representation tasks. For example, the dynamics-aware strategy effectively detects the dynamic components in the 3D space with available knowledge. This capability particularly enhances applications such as telepresence and holographic communication, where DASS can selectively transmit and stream only the dynamic or human-related components. Consequently, this approach enhances the computational efficiency and reduces communication overhead, which are critical requirements in current 2D online meeting applications. Besides, the error-guided densification proves effective in rapidly compensating for areas with weak reconstruction. It is potential to be extended to further applications where Gaussian densification is performed. 

In the aspect of performance improvement, similar to other iterative streamable 4D reconstruction methods, our DASS consistently benefits from the prospective higher quality of initialization, faster rasterization, and improved training strategies in the 3DGS field, as well as frame-wise and batch-wise parallel computation. 

For future extensions of DASS, we plan to consider two key areas: fast on-the-fly training under sparse viewpoint inputs and adaptation to significant scene changes, which have been long-standing challenges in spatial representation tasks. These scenarios also complicate the on-the-fly training task, which relies only on previous information and current multi-view inputs.

% \clearpage

\begin{table*}
    \centering
    % \resizebox{\columnwidth}{!}{%
    \begin{tabular}{@{}l|c|c|c|c|c|c|c@{}}
      \toprule
      % \hline
      \multirow{2}{*}{Method} & {Coffee} & {Cook} & {Cut} & {Flame} & {Flame} & {Sear} & \multirow{2}{*}{Mean} \\
                              & {Martini} & {Spinach} & {Beef} & {Salmon} & {Steak} & {Steak} &  \\
      \midrule
      % \hline
      Plenoxels~\cite{fridovich2022plenoxels}  & 27.65 & 31.73 & 32.01 & 28.68 & 32.24 & 32.33 & 30.77 \\
      I-NGP~\cite{muller2022instant}      & 25.19 & {29.84} & 30.73 & 25.51 & {30.04} & 30.40 & {28.62} \\
      \midrule
      % \hline
      DyNeRF$^\dagger$~\cite{li2022neural}  & {--} & {--} & {--} & 29.58 & {--} & {--} & 29.58 \\
      NeRFPlayer~\cite{song2023nerfplayer} & \cellcolor{rank1}31.53 & 30.58 & 29.35 & \cellcolor{rank1}31.65 & 31.93 & 29.13 & 30.69 \\
      HexPlane~\cite{cao2023hexplane}   & {--} & 32.04 & 32.55 & 29.47 & 32.08 & 32.39 & 31.70 \\
      K-Planes~\cite{fridovich2023k}   & \cellcolor{rank2}29.99 & 32.60 & 31.82 & \cellcolor{rank2}30.44 & 32.38 & 32.52 & 31.63 \\
      HyperReel~\cite{attal2023hyperreel}  & 28.37 & 32.30 & 32.92 & 28.26 & 32.20 & 32.57 & 31.10 \\
      MixVoxels~\cite{wang2023mixed} & \cellcolor{rank3}29.36 & 31.61 & 31.30 & \cellcolor{rank3}29.92 & 31.43 & 31.21 & 30.80 \\
      \midrule
      4DGS-Wu~\cite{wu20244d} & 27.34 & 32.46 & 32.90 & 29.20 & 32.51 & 32.49 & 31.15 \\
      4DGS-Yang~\cite{yangreal} & 28.33 & 32.93 & \cellcolor{rank1}33.85 & 29.38 & \cellcolor{rank3}34.03 & \cellcolor{rank2}33.51 & \cellcolor{rank2}32.01 \\
      E-D3DGS~\cite{bae2024per} & 29.10 & 32.96 & \cellcolor{rank2}33.57 & 29.61 & 33.57 & \cellcolor{rank3}33.45 & 31.31 \\
      STG$^\ddagger$~\cite{li2024spacetime} & 28.61 & \cellcolor{rank3}33.18 & 33.52 & 29.48 & 33.64 & \cellcolor{rank1}33.89 & \cellcolor{rank1}32.05 \\
      \midrule
      StreamRF~\cite{li2022streaming}   & 27.88 & 31.55 & 31.90 & 28.31 & 32.28 & 32.33 & 30.71 \\
      3DGStream~\cite{sun20243dgstream}  & 27.69 & \cellcolor{rank2}33.39 & 33.11 & 28.45 & \cellcolor{rank1}34.30 & 33.16 & 31.68 \\
      Ours & 28.15 & \cellcolor{rank1}33.83 & \cellcolor{rank3}33.54 & 28.84 & \cellcolor{rank2}34.26 & 33.33 & \cellcolor{rank3}31.99 \\
      \bottomrule
    \end{tabular}
    % }
    \caption{\textbf{Quantitative comparison} of PSNR values across all scenes in the N3DV dataset, with the metric for each scene calculated as the average over 300 frames. 
    $^\dagger$DyNeRF~\cite{li2022neural} only reports metrics on the \textit{flame salmon} scene.
    $^\ddagger$STG~\cite{li2024spacetime} trains each model with a 50-frame video sequence, requiring six models to complete the overall representation.
    }
    \label{tab:N3DV_Comparisons}
\end{table*}
% \vspace{10mm}

\newpage

\end{document}